\documentclass[twoside]{article}

\usepackage[accepted]{aistats2021}
\usepackage{graphicx}
\usepackage{natbib}
\usepackage{algorithm}
\usepackage[noend]{algpseudocode}
\usepackage{algorithmicx}
\usepackage[dvipsnames]{xcolor}
\usepackage{amsthm}
\usepackage{multirow} 
\usepackage{breakcites}

\usepackage{hyperref}
\hypersetup{
    colorlinks=true,
    linkcolor=blue,
    filecolor=magenta,
    citecolor=black,
    urlcolor=cyan,
}

 \usepackage{listings}
\usepackage{multicol}
\usepackage{multirow} 
\usepackage{ulem}
\usepackage{hyperref}
\definecolor[named]{DarkBlue}{cmyk}{1,1,0,0.45}
\definecolor[named]{Sepia}{cmyk}{0,0.411,0.821,0.561}
\definecolor[named]{DarkOrchid}{cmyk}{0.25,0.75,0,0.20}
\definecolor[named]{DarkGreen}{cmyk}{1,0,1,0.61}

\lstdefinestyle{PClean}{
  basicstyle=\ttfamily\scriptsize,
  columns=fullflexible,
  keepspaces=true,
  alsoletter={\.,\%,\#, \@, \?, \/, \!},
  morekeywords=[1]{function, subproblem, class, latent, parameter,preferring, index,on, observation, if, else, elseif, end, for, begin, in, const, struct, using, return, const, do, quote},
  keywordstyle=[1]\ttfamily\bf,
  morekeywords=[2]{\@static, \@class, \@pclean, \@column, \@dirty, \@gen, \@inj, \@trace, \@param, \@input, \@output,
                  \@tf_function, \@call, \@read, \@tf_call, \@copy,
                  \@splice, \@write,\@choicediff,\@diff,\@ad,\@tensorflow_function},
  keywordstyle=[2]\textcolor{DarkBlue},
  morekeywords=[3]{clean,fit!,initialize, generate, choicemap, map_optimize, importance_resampling, select, mala, mh, metropolis_hastings, update, propose, assess, custom_importance, train_batched!, TFFunction},
  keywordstyle=[3]\textcolor{NavyBlue},
  morekeywords=[4]{add_noise,add_typos,time_prior,maybe_swap,choose_proportionally,GPT2_elaborate,normal,bernoulli,gamma,transformed_gaussian, unmodeled,dirichlet,uniform,beta,independent_pixel_noise,string_prior,discrete,typos},
  keywordstyle=[4]\textcolor{DarkGreen},
  morekeywords=[5]{},
  keywordstyle=[5]\textcolor{BrickRed},
  showstringspaces=False,
  stringstyle=\ttfamily\color{DarkOrchid},
  morestring=[b]{"},
  morecomment=[l]{\#},
  commentstyle=\color{darkgray}\ttfamily,
  escapeinside={<@}{@>},
  numbers=left,
  numberstyle=\tiny\color{darkgray},
  numbersep=3pt
}

\newcommand\blfootnote[1]{%
  \begingroup
  \renewcommand\thefootnote{}\footnote{#1}%
  \addtocounter{footnote}{-1}%
  \endgroup
}

\usepackage[symbol]{footmisc}

\renewcommand{\thefootnote}{\fnsymbol{footnote}}

\begin{document}

\algblockdefx[model]{Model}{EndModel}%
[2]{\textsc{#1}(#2):}{}
\algnewcommand{\LineComment}[1]{\State \(\triangleright\) #1}

%

%

\twocolumn[
\runningtitle{PClean: Bayesian Data Cleaning at Scale with Domain-Specific Probabilistic Programming}
\aistatstitle{PClean: Bayesian Data Cleaning at Scale with\\ Domain-Specific Probabilistic Programming}

\aistatsauthor{ Alexander K. Lew \And Monica Agrawal \And  David Sontag \And Vikash K. Mansinghka} \aistatsaddress{Massachusetts Institute of Technology} ]

\begin{abstract}
Data cleaning is naturally framed as probabilistic inference in a generative model of ground-truth data and likely errors, but the diversity of real-world error patterns and the hardness of inference make Bayesian approaches difficult to automate. 
We present PClean, a probabilistic programming language (PPL) for leveraging dataset-specific knowledge to automate Bayesian cleaning.\blfootnote{All code available at \href{https://github.com/probcomp/PClean}{github.com/probcomp/PClean}} 
Compared to general-purpose PPLs, PClean tackles a restricted problem domain, enabling three modeling and inference innovations: (1) a non-parametric model of relational database instances, which users' programs customize; (2) a novel sequential Monte Carlo inference algorithm that exploits the structure of PClean's model class; and (3) a compiler that generates near-optimal SMC proposals and blocked-Gibbs rejuvenation kernels based on the user's model and data.
We show empirically that short (\textless 50-line) PClean programs can: be faster and more accurate than generic PPL inference on data-cleaning benchmarks; match state-of-the-art data-cleaning systems in terms of accuracy and runtime (unlike generic PPL inference in the same runtime); and scale to real-world datasets with millions of records.

\end{abstract}

\section{INTRODUCTION}

\begin{figure*}
    \centering
    \includegraphics[width=0.9\textwidth]{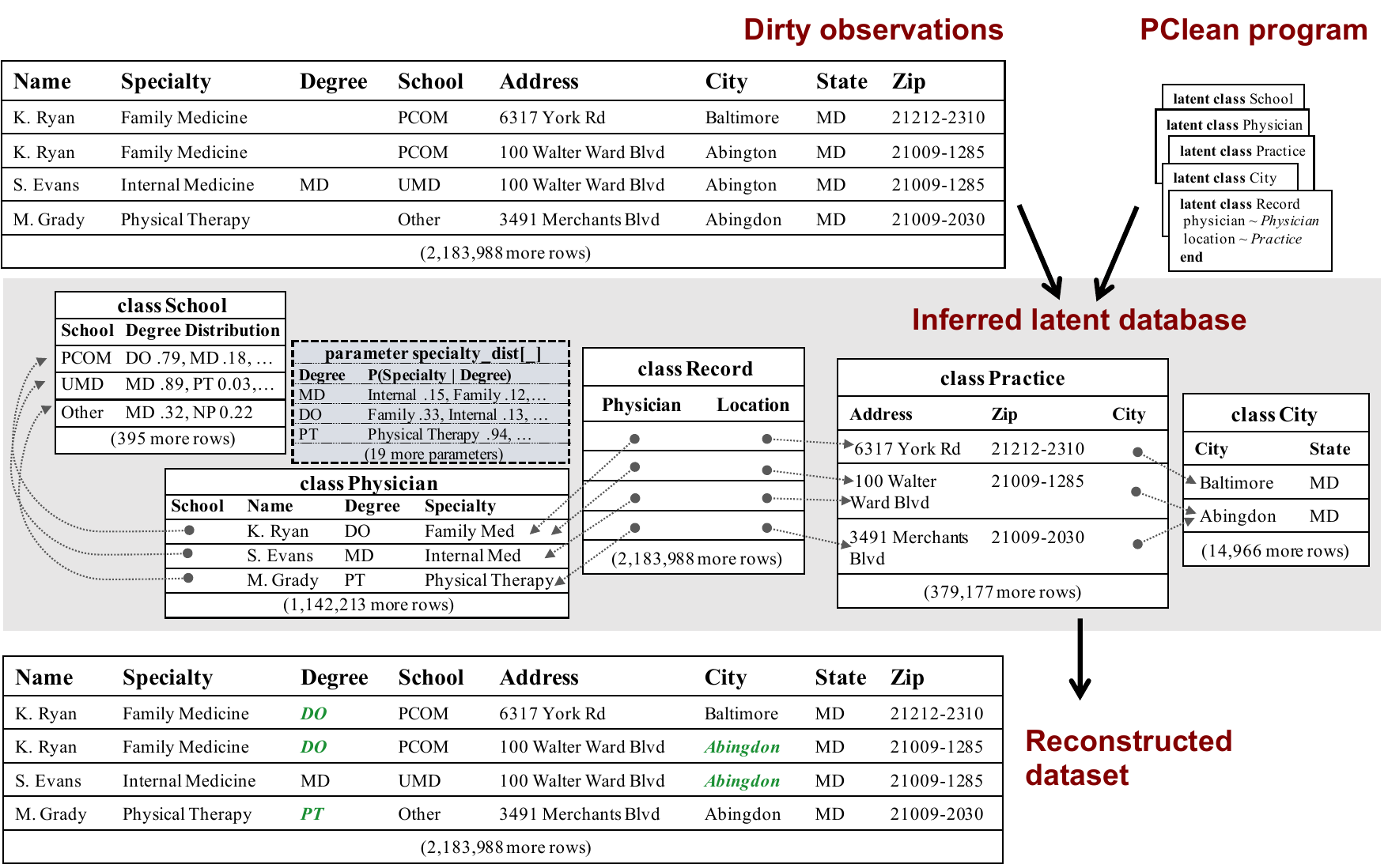}
    \caption{PClean applied to  Medicare's 2.2-million-row Physician Compare National database. Based on a user-specified relational model, PClean infers a latent database of entities, which it uses to correct systematic errors (e.g. the misspelled \textit{Abington, MD} appears 152 times in the dataset) and impute missing values.}
    \vspace{-1mm}
    \label{fig:physicians_results}
\end{figure*}

Real-world data is often noisy and incomplete, littered with NULLs, typos, duplicates, and inconsistencies. Cleaning dirty data is important for many workflows, but can be difficult to automate, requiring judgment calls about objects in the world (e.g., to decide whether two records refer to the same hospital, or which of several cities called ``Jefferson'' someone lives in). Generative models provide a conceptually appealing approach to automating this sort of reasoning, but the diversity of real-world errors~\citep{Abedjan2016} and the difficulty of inference pose significant challenges.

This paper presents PClean, a domain-specific generative probabilistic programming language (PPL) for Bayesian data cleaning.
 As in some existing PPLs (e.g. BLOG~\citep{Milch2006}), PClean programs encode prior knowledge about relational domains, and quantify uncertainty about the latent networks of objects that underlie observed data. However, PClean's approach is inspired by {\it domain-specific} PPLs, such as Stan \citep{Carpenter2017} and Picture \citep{Kulkarni2015}: it aims not to serve all conceivable relational modeling needs, but rather to deliver fast inference, concise model specification, and accurate cleaning on large-scale problems. It does this via three modeling and inference contributions:
\vspace{-2.5mm}
\begin{enumerate}
    \item PClean introduces a domain-general non-parametric  prior on the number of latent objects and their link structure.  PClean programs customize the prior via a relational schema and via generative models for objects' attributes.
    \vspace{-2mm}
    \item PClean inference is based on a novel sequential Monte Carlo (SMC) algorithm, to initialize a latent object database with plausible guesses, and novel rejuvenation updates to fix mistakes.
    \vspace{-2mm}
    \item PClean provides a compiler that generates near-optimal SMC proposals and Gibbs rejuvenation kernels given the user’s data, PClean program, and inference hints. These proposals improve over generic top-down PPL inference by incorporating local Bayesian reasoning within user-specified subproblems, and heuristics from traditional cleaning systems.
    
\end{enumerate}
    \vspace{-2mm}
Together, these innovations improve over generic PPL inference, enabling fast and accurate cleaning of challenging real-world datasets with millions of rows.


\subsection{Related Work}

Many researchers have proposed  generative models for cleaning specific datasets or fixing particular error patterns~\citep{Pasula2003, Kubica2003,  MayfieldJenniferNeville2009, Matsakis2010, Xiong2011, Hu2012,  Zhao2012, Abedjan2016,  De2016, Steorts2016, Winn2017, DeSa2019, eduardo2020robust,marchant2021d}. Such formulations specify priors over ground truth data, and likelihoods that model errors. In contrast, PClean's PPL makes it easy to write short (\textless 50 line) programs to specify custom models of new datasets, and automates an inference algorithm that delivers fast, accurate cleaning results. 


PClean draws on a rich literature of Bayesian approaches to modeling relational data~\citep{Friedman1999}, including open-universe models with identity and existence uncertainty~\citep{Milch2006}. Many PPLs could express PClean-like data cleaning models \citep{milch2005, goodman2012church, dippl, Mansinghka2014, Gordon2014, Tolpin2016, scibior2018functional,  bingham2019, Cusumano-Towner2019}, but in practice, generic PPL inference is often too slow. This paper introduces new algorithms that 
scale better, and demonstrates external validity of the results by calibrating PClean's runtime and accuracy against SOTA data-cleaning baselines~\citep{Dallachiesat2013,Rekatsinas2017} that use machine learning and weighted logic (typical of discriminative approaches~\citep{Mccallum2003, Wellner2004, Wick2013}). 

Some of PClean's inference innovations have close analogues in traditional  cleaning systems; for example, PClean's preferred values from Section~\ref{sec:hints} are related to HoloClean's notion of domain restriction. In fact, PClean can be viewed as a scalable, Bayesian, domain-specific PPL implementation of the PUD framework from \citet{DeSa2019} (which abstractly characterizes the HoloClean implementation from~\citet{Rekatsinas2017}, but does not itself include PClean's modeling or inference innovations). Some of PClean's inference contributions also have precursors in LibBi and Birch~\citep{Murray2015,murray2018automated}, which, like PClean, employ sequential Monte Carlo algorithms with data-driven proposals. However, Birch's delayed sampling technique~\citep{murray2018delayed} would {\it not} yield intelligent, data-driven proposals in PClean's non-parametric model class, and in Section~\ref{sec:results}, we show that PClean's novel inference contributions (including its static generation of model-specific proposal code, and its per-object rejuvenation schedule) are necessary for efficient, accurate cleaning.

\section{MODELING}
\label{sec:modeling}
In this section, we present the PClean modeling language, which is designed for the concise encoding of domain-specific knowledge about data and likely errors into generative models. PClean programs specify (i) a prior $p(\mathbf{R})$ over a latent ground-truth relational database of entities, and (ii) an observation model $p(\mathbf{D} \mid \mathbf{R})$ describing how the attributes of entities from $\mathbf{R}$ are reflected in an observed flat data table $\mathbf{D}$. Unlike general-purpose PPLs, PClean does not afford complete freedom in specifying $p(\mathbf{R})$. Instead, we impose a novel domain-general structure prior $p(\mathbf{S})$ on the \textit{skeleton} of the database $\mathbf{R}$: $\mathbf{S}$ determines how many entities are in each latent database table, and which entities are related. The user's program encodes only $p(\mathbf{R} \mid \mathbf{S})$, a probabilistic relational model over the attributes of the objects whose existence and relationships are given by $\mathbf{S}$. This decomposition limits the PClean model class, but enables the development of an efficient SMC inference algorithm (Section~\ref{sec:inference}).

\subsection{PClean Modeling Language}
\label{sec:language}
A PClean program defines a set of \textit{classes}
$\mathcal{C} = (C_1, \dots, C_k)$, one for each type of object 
underlying the user's data, and a \textit{query} $\mathbf{Q}$ that describes how latent objects inform the observed flat dataset $\mathbf{D}$.

\begin{figure}[t]
\includegraphics[width=\linewidth]{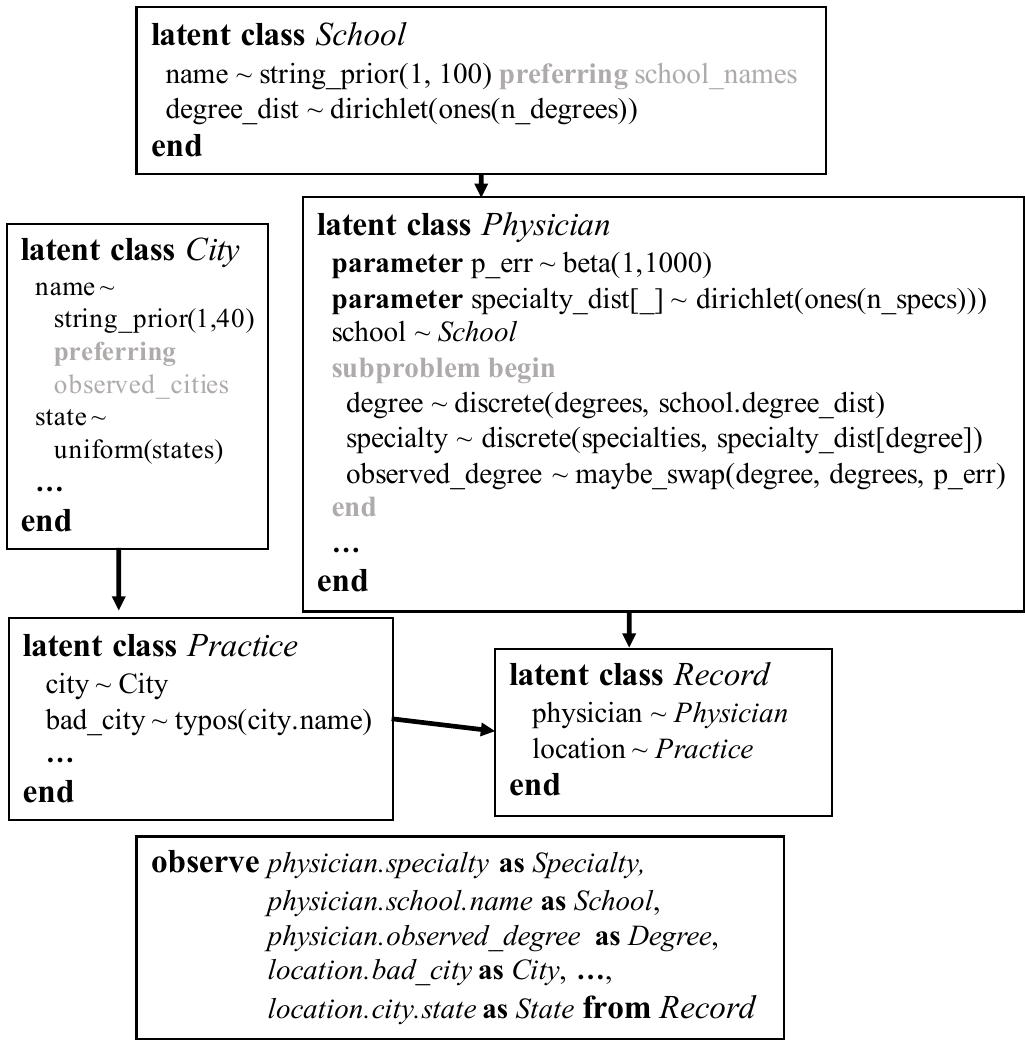}
\caption{An example PClean program. PClean programs define: (i) an \textit{acyclic} relational schema, comprising a set of classes $\mathcal{C}$, and for each class $C$, sets $\mathcal{A}(C)$ of attributes and $\mathcal{R}(C)$ of reference slots; (ii) a probabilistic relational model $\Pi$ encoding priors for object attributes; and (iii) a query $\mathbf{Q}$ (last line), specifying how attributes are observed in the flat data table $\mathbf{D}$. \textit{Inference hints} (in gray) do not change the model.}
\label{fig:example-program}
\vspace{-4mm}
\end{figure}

\textbf{Class Declarations.} The declaration of a PClean class $C$ includes
three kinds of statement: \textit{reference statements} ($Y \sim C'$), which define a  reference slot $C.Y$ connecting objects of class $C$ to objects of a target class $T(C.Y) = C'$; \textit{attribute statements} ($X \sim \phi_{C.X}(\dots)$), which define a new \textit{attribute} $C.X$ that objects of the class possess, and declare the prior distribution $\phi_{C.X}$ that it follows; and $\textit{parameter statements}$ ($\textbf{parameter } \theta_C \sim p_{\theta_C}(\dots)$), which introduce mutually independent hyperparameters shared by all objects of the class $C$, to be learned from the noisy data. The prior $\phi_{C.X}$ for an attribute may depend on the values of a \textit{parent set} $Pa(C.X)$ of attributes, potentially accessed via reference slots. For example, in Figure~\ref{fig:example-program}, the \textit{Physician} class has a \textit{school} reference slot with target class \textbf{School}, and a \textit{degree} attribute whose value depends on \textit{school.degree\_dist}. Together, the attribute statements specify a \textit{probabilistic relational model} $\Pi$ for the user's schema (possibly parameterized by hyperparameters $\{\theta_C\}_{C \in \mathcal{C}}$)~\citep{Friedman1999}.

\textbf{Query.} A PClean program ends with a \textit{query}, connecting the schema of the latent relational database to the fields of the observed dataset. The query has the form \textbf{observe} $(U_1 \textbf{ as } x_1), \cdots, (U_k \textbf{ as } x_k) \textbf{ from } C_{obs}$, where $C_{obs}$ is a class that models the records of the observed dataset (\textit{Record}, in Figure~\ref{fig:example-program}), $x_i$ are the names of the columns in the observed dataset, and $U_i$ are dot-expressions (e.g., \textit{physician.school.name}), picking out attributes accessible via zero or more reference slots from $C_{obs}$. 
The records of the dataset $\mathbf{D}$ are modeled as directly recording the values of the attributes named by $U_i$, each for a distinct object in $C_{\textit{obs}}$.
As such, errors must be modeled as \textit{part} of the latent database $\mathbf{R}$, not as a separate stage of the generative process. For example, Figure~\ref{fig:example-program} models systematic typos in the \textit{City} field, by associating each \textit{Practice} with a possibly misspelled version \textit{bad\_city} of the practice's city.

\begin{figure}[t!]
    \centering
    \includegraphics[width=\linewidth]{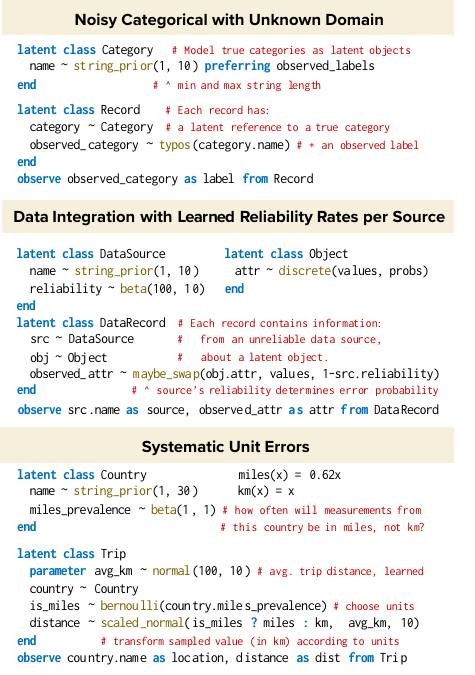}
    \vspace{-4mm}
    \caption{PClean programs can concisely model a variety of data-cleaning scenarios.}
    \label{fig:examples}
    \vspace{-4mm}
\end{figure}

We emphasize that the relational schema and query are modeling choices: much of PClean's expressive power comes from the freedom to posit latent relational structure that is not directly reflected in the dataset. Figure~\ref{fig:examples} shows how this freedom can be used to capture several common data-cleaning motifs.

\subsection{Non-Parametric Structure Prior $p(\mathbf{S})$}

\begin{figure}
\begin{algorithmic}[0]
\Model{GenerateSkeleton}{$\mathcal{C}, |\mathbf{D}|$}
\LineComment{Create one $C_{obs}$ object per observed record}
\State $\mathbf{S}_{C_{obs}} := \{1, \dots, |\mathbf{D}|\}$
\LineComment{Generate a class \textit{after} all referring classes:}
\For{class $C \in \textsc{TopoSort}(\mathcal{C} \setminus \{C_{obs}\})$}
  \LineComment{Collect references to class $C$}
  \State $\mathbf{Ref}_\mathbf{S}^C := \{(r, Y) \mid r \in \mathbf{S}_{C'}, T(C'.Y) = C\}$
  \LineComment{Generate targets of those references}
  \State $\mathbf{S}_C \sim \textsc{GenerateObjectSet}(C, \mathbf{Ref}_\mathbf{S}^C)$
  \LineComment{Assign reference slots pointing to $C$}
  \For{object $r' \in \mathbf{S}_C$}
  	\For{referring object $(r, Y) \in r'$}
  		\State $r.Y := r'$
  	\EndFor
  \EndFor
\EndFor
\LineComment{Return the skeleton}
\State \Return $\{\mathbf{S}_C\}_{C \in \mathcal{C}}, (r, Y) \mapsto r.Y$
\EndModel

\Model{GenerateObjectSet}{$C, \mathbf{Ref}_{\mathbf{S}}^C$}
\State $s_C \sim \textit{Gamma}(1, 1)$; $d_C \sim \textit{Beta}(1, 1)$ 
\LineComment{\parbox[t]{.9\linewidth}{Partition references into co-referring subsets}}
\State $\mathbf{S}_C \sim CRP(\mathbf{Ref}_{\mathbf{S}}^C, s_C, d_C)$
\State \Return $\mathbf{S}_C$
\EndModel
\end{algorithmic}

\caption{PClean's non-parametric structure prior $p(\mathbf{S})$ over the relational skeleton $\mathbf{S}$ for a schema $\mathcal{C}$.}
\label{fig:skeleton-generator}
\vspace{-3mm}
\end{figure}

A PClean program's class declarations specify a probabilistic relational model that can be used to generate the attributes of objects in the latent database, but does not encode a prior over how many objects exist in each class or over their relationships. (The one exception is $C_{obs}$, the designated observation class, whose objects are assumed to be in one-to-one correspondence with the rows of the observed dataset $\mathbf{D}$.) In this section, we introduce a domain-general structure prior $p(\mathbf{S}; |\mathbf{D}|, \mathcal{C})$ that encodes a non-parametric generative process over the \textit{object sets} $\mathbf{S}_C$ associated with each class $C$, and over the values of each object's reference slots. The parameter $|\mathbf{D}|$ is the number of observed data records; $p(\mathbf{S}; |\mathbf{D}|, \mathcal{C})$ places mass only on relational skeletons in which there are exactly $|\mathbf{D}|$ objects in $C_{obs}$ and every other object is connected via some chain of reference slots to one of them.

PClean's generative process for relational skeletons is shown in Figure~\ref{fig:skeleton-generator}. First, with probability 1, we set $\mathbf{S}_{C_{obs}} = \{1, \dots, |\mathbf{D}|\}$ (a set of $|\mathbf{D}|$ distinct object IDs). PClean requires that the directed graph with edges $(C, T(C.Y))$ for each reference slot $C.Y$ be acyclic, which allows us to generate the remaining object sets class-by-class, processing a class only {\it after} any classes with reference slots targeting it.
To generate an object set for class $C$, we first consider the reference set $\mathbf{Ref}_\mathbf{S}^C$ of all objects with reference slots targeting $C$:
$$\mathbf{Ref}_\mathbf{S}^C = \{(r, Y) \mid Y \in \mathcal{R}(C') \wedge T(C'.Y) = C\wedge r \in \mathbf{S}_{C'}\}$$
The elements of $\mathbf{Ref}_\mathbf{S}^C$ are pairs $(r, Y)$ of an object and a reference slot; if a single object has two reference slots targeting class $C$, then the object will appear twice in the reference set. The point is to capture all of the places in $\mathbf{S}$ that will refer to objects of class $C$.

We then generate a \textit{co-reference partition} of $\mathbf{Ref}_\mathbf{S}^C$, i.e., we partition the references to class $C$ into disjoint subsets, within each of which we take all references to target the same object. To do this, we use the two-parameter Chinese restaurant process $CRP(X, s, d)$, which defines a non-parametric distribution over partitions of its set-valued parameter $X$. The strength $s$ and discount $d$ control the sizes of the clusters.
The CRP generates a partition of all references to class $C$, and \textit{we treat the resulting partition as the object set $\mathbf{S}_C$}, i.e., each component defines one object of class $C$:
$$\mathbf{S}_C \mid \mathbf{Ref}_\mathbf{S}^C \sim CRP(\mathbf{Ref}_{\mathbf{S}}^C, s_C, d_C)$$ 
To set the reference slots $r.Y$ with target class $T(\mathbf{Class}(r).Y) = C$, we simply look up which partition component $(r, Y)$ (viewed as an element of $\mathbf{Ref}_{\mathbf{S}}^C$) was assigned to. Since we have equated these partition components with objects of class $C$, we can directly set $r.Y$ to point to the component (object) that contains $(r, Y)$ as an element:
$$r.Y := \textrm{the unique } r' \in \mathbf{S}_{T(\mathbf{Class}(r).Y)} \textrm{ s.t. } (r, Y) \in r'$$
This procedure can be applied iteratively to generate object sets for every class and fill all reference slots.

\section{INFERENCE}
\label{sec:inference}

PClean's non-parametric structure prior ensures that PClean models admit a sequential
representation, which can be used as the basis of a resample-move sequential Monte Carlo inference scheme
 (Section~\ref{sec:smc}). However, if the SMC and rejuvenation proposals are made
from the model prior, as is typical in PPLs, inference will still require
prohibitively many particles to deliver accurate results. To address this issue, PClean uses
a \textit{proposal compiler} that exploits conditional independence in the model
to generate fast enumeration-based proposal kernels for both SMC and MCMC rejuvenation (Section~\ref{sec:proposals}). Finally, to help users scale these proposals to large data,
we introduce \textit{inference hints}, lightweight annotations in the PClean program that
can divide variables into subproblems to be separately handled by the proposal, or direct
the enumerator to focus its efforts on a dynamically computed subset of a large discrete domain (Section~\ref{sec:hints}).

\subsection{Per-Observation Sequential Monte Carlo with Per-Object Rejuvenation}
\label{sec:smc}

One version of the PClean model's generative process was given in Section~\ref{sec:modeling}: a skeleton can be generated from $p(\mathbf{S})$, then attributes can be filled in using the user's probabilistic relational model $p_{\Pi}(\mathbf{R} \mid \mathbf{S})$. Finally an observed dataset $\mathbf{D}$ can be generated according to the query $\mathbf{Q}$. But importantly, the model also admits a sequential representation, in which the latent database $\mathbf{R}$ is built in stages: at each stage, a single record is
added to the observation class $C_{obs}$, along with any new objects in other classes that it refers to. Using this representation, we can run SMC on the model, building a particle approximation to the posterior that incorporates one observation at a time.

\textbf{Database Increments.} Let $\mathbf{R}$ be a database with designated observation class $C_{obs}$. Assume $\mathbf{R}_{C_{obs}}$, the object set for the class $C_{obs}$, is $\{1, \dots, |\mathbf{D}|\}$. Then the database's $i^{\textrm{th}}$ \textit{increment} $\Delta_\mathbf{R}^i$ is the object set
$$\{r \in \mathbf{R} \mid \exists K, \, i.K = r \wedge \forall K', \forall j < i, j.K' \neq r\},$$
along with their attribute values and targets of their reference slots. Objects in $\Delta_\mathbf{R}^i$ 
may refer to other objects within the increment, or in earlier increments.
Intuitively, the $i^{\textrm{th}}$ increment of a database is the set of objects referenced by the $i^{\textrm{th}}$ observation object, but \textit{not} by any previous observation object $j < i$.

\textbf{Sequential Generative Process.} Figure~\ref{fig:seqrep} shows a generative process equivalent to the one in Section~\ref{sec:modeling}, but which generates the attributes and reference slots of each increment sequentially. Intuitively, the database is generated via a Chinese-restaurant `social network':  Consider a collection of restaurants, one for each class $C$, where each table serves a dish $r$ representing an object of class $C$.
Upon entering a restaurant, customers either sit at an existing
table or start a new one, as in the usual generalized CRP construction.
But these restaurants require that to start a new table, customers must first send $
|\mathcal{R}(C)|$ friends to \textit{other} restaurants (one to the target of each reference slot). Once the friends are seated at these \textit{parent} restaurants, the original customer samples attributes $r.X$ of the new table's object, possibly informed by \textit{their friends'} dishes (the objects $r.Y$ of class $T(C.Y)$). 
The process starts with $|\mathbf{D}|$ customers at the restaurant for $C_{Obs}$,
who sit at separate tables; each customer who sits down triggers the sampling of one increment.

\begin{figure}
\begin{algorithmic}[0]
\Model{GenerateDataset}{$\Pi$, $\mathbf{Q}$, $|\mathbf{D}|$}
\State $\mathbf{R}^{(0)} \leftarrow \emptyset$ \Comment{Initialize empty database}
\For{observation $i \in \{1, \dots, |\mathbf{D}|\}$}
  
  \State $\Delta_i^\mathbf{R} \leftarrow \textsc{GenerateDbIncr}(\mathbf{R}^{(i-1)}, C_{obs})$
  \State $\mathbf{R}^{(i)} \leftarrow \mathbf{R}^{(i-1)} \cup \Delta_i^\mathbf{R}$

  \State $r \leftarrow$ the unique object of class $C_{obs}$ in $\Delta_i^\mathbf{R}$
  \State $d_i \leftarrow \{X \mapsto r.\mathbf{Q}(X),\,\, \forall X \in \mathcal{A}(\mathbf{D})\}$
\EndFor
\State \Return $\mathbf{R} = \mathbf{R}^{(|\mathbf{D}|)}, \mathbf{D} = (d_1, \dots, d_{|\mathbf{D}|})$
\EndModel
\Model{GenerateDbIncr}{$\mathbf{R}^{(i-1)}$, root class $C$}
  \State $\Delta \leftarrow \emptyset$; $r_* \leftarrow$ a new object of class $C$
  \For{each reference slot $Y \in \mathcal{R}(C)$}
  	\State $C' \leftarrow T(C.Y)$
  	\For{each object $r \in \mathbf{R}^{(i-1)}_{C'} \cup \Delta_{\mathbf{R}_{C'}}$}
  		\State $n_r \leftarrow |\{r'\mid r' \in \mathbf{R}^{(i-1)} \cup \Delta \wedge \exists \tau, r'.\tau = r\}|$
  	\EndFor
  	\State $r_*.Y \leftarrow r$ w.p. $\propto {n_r - d_{C'}}$, or
  	\State \,\,\,\,\,\,\,\,\,\,\,\,\,\,\,\, \,\,\,$\star$ w.p. $\propto {s_{C'} + d_{C'}|\mathbf{R}^{(i-1)}_{C'} \cup \Delta_{\mathbf{R}_{C'}}|}$
  	\If{$r_*.Y = \star$}
  		\State $\Delta' \leftarrow \textsc{GenerateDbIncr}(\mathbf{R}^{(i-1)} \cup \Delta, C')$
  		\State $\Delta \leftarrow \Delta \cup \Delta'$
  		\State $r_*.Y \leftarrow $ the unique $r'$ of class $C'$ in $\Delta'$
  	\EndIf
  \EndFor
  \For{each $X \in \mathcal{A}(C)$, in topological order}
  	\State $r_*.X \sim \phi_{C.X}(\cdot \mid \{r_*.U\}_{U \in Pa(C.X)})$
  \EndFor
  \State \Return $\Delta \cup \{r_*\}$
\EndModel

\end{algorithmic}
\vspace{-4mm}
\caption{Sequential model representation.}
\label{fig:seqrep}
\end{figure}

\begin{algorithm*}[th!]
\caption{Compiling SMC proposal to Bayesian network}
\label{alg:increment-bayes-net}
\begin{algorithmic}
\Procedure{GenerateIncrementBayesNet}{partial instance $\mathbf{R}^{(i-1)}$, data $d_i$}
  \LineComment{Set the vertices to all attributes and reference slots accessible from $C_{obs}$}
  \State $U \leftarrow \mathcal{A}(C_{obs}) \cup \{K \mid C_{obs}.K \textrm{ is a valid slot chain} \} \cup \{K.X \mid X \in \mathcal{A}(T(C_{obs}.K))\}$ 
  \LineComment{Determine parent sets and CPDs for each variable}
  \For{each variable $u \in U$}
  	\If{$u \in \mathcal{A}(C_{obs})$}
  		\State Set $Pa(u) = Pa^{\Pi}(C.u)$
  		\State Set $\phi_{u}(v_u \mid \{v_{u'}\}_{u' \in Pa(u)}) = \phi^\Pi_{C.u}(v_u \mid \{v_{u'}\}_{u' \in Pa(u)})$
  	\ElsIf{$u = K.X$ for $X \in \mathcal{A}(T(C_{obs}.K))$}
  		\State Set $Pa(u) = Pa^{\Pi}(T(C_{obs}.K).X) \cup \{K\} \cup \{u'.X \mid u' \textrm{ already processed} \wedge T(C_{obs}.u') = T(C_{obs}.K)\}$
  		\State Set \[ \phi_u(v_u \mid \{v_{u'}\}_{u' \in Pa(u)}) = \begin{cases} 
      \mathbf{1}[v_u = v_K.X] & v_K \in \mathbf{R}^{(i-1)} \\
      \phi^\Pi_{T(C_{obs}.K).X}(v_u \mid \{v_{u'}\}_{u' \in Pa^\Pi(T(C_{obs}.K).X)}) & v_K = \textbf{new}_K \\
      \mathbf{1}[v_u = v_{u'.X}] & v_K = \textbf{new}_{u'}, u' \neq K
   \end{cases}
\]
  	\Else
  	    \State Set $Pa(u)$ to already-processed slot chains $u'$ s.t. $T(C.u') = T(C.u)$, and $K$ if $u = K.Y$
  	    \State Set domain $V(u) = \mathbf{R}^{(i-1)}_{T(C.u)} \cup \{\textbf{new}_{u'} \mid u' \in Pa(u) \cup \{u\}\}$
  	    \State Set $\phi_u(v_u \mid \{v_{u'}\}_{u' \in Pa(u)})$ according to CRP, or to $\mathbf{1}[v_u=v_K.Y]$ if $u = K.Y$ and $v_K \in \mathbf{R}^{(i-1)}$
  	\EndIf
  \EndFor

  \For{attribute $X \in \mathcal{A}(\mathbf{D})$}
  	\State Change node $\mathbf{Q}(u)$ to be observed with value $d_i.x$, \textbf{unless} $d_i.x$ is missing
  \EndFor
\EndProcedure
\end{algorithmic}
\end{algorithm*}

\textbf{SMC Inference with Per-Object Rejuvenation.}
The sequential representation yields a sequence of intermediate unnormalized target densities $\tilde{\pi}_i$ for SMC:
$$\tilde{\pi}_i(\mathbf{R}) = \prod_{j=1}^i p(\Delta_j^\mathbf{R} \mid \Delta_1^\mathbf{R}, \dots, \Delta_{j-1}^\mathbf{R}) p(d_j \mid \Delta_1^\mathbf{R}, \dots, \Delta_j^\mathbf{R}).$$
Particles are initialized to hold an empty database, to which proposed increments $\Delta_i^\mathbf{R}$ are added each iteration. As is typical in SMC, at each step, the particles are reweighted according to how well they explain the new observed data, and resampled to cull low-weight particles while cloning and propagating promising ones. This process allows the algorithm to hypothesize new latent objects as needed to explain each new observation, but not to revise earlier inferences about latent objects (or delete previously hypothesized objects) in light of new observations; we address this problem with MCMC rejuvenation moves. These moves select an object $r$, and update all $r$'s attributes and reference slots in light of all relevant data incorporated so far. In doing so, these moves may also lead to the ``garbage collection'' of objects that are no longer connected to the observed dataset, or to the insertion of new objects as targets of $r$'s reference slots. 
\vspace{-3mm}

\subsection{Compiling Data-Driven SMC Proposals}
\label{sec:proposals}
Proposal quality is the determining factor for the quality of SMC inference: at each step of the algorithm, a proposal $Q_i(\Delta_i^\mathbf{R}; \mathbf{R}^{(i-1)}, d_i)$ generates proposed additions $\Delta_i^\mathbf{R}$ to the existing latent database $\mathbf{R}^{(i-1)}$ to explain the $i^\textrm{th}$ observed data point, $d_i$. A key limitation of the sequential Monte Carlo implementations in most general-purpose PPLs today is that the proposals $Q_i$ are not \textit{data-driven}, but rather based only on the prior: they make blind guesses as to the latent variable values and thus tend to make proposals that explain the data poorly. 
By contrast, PClean compiles proposals that use exact enumerative inference to propose discrete variables in a data-driven way. This approach extends ideas from \citet{arora2012gibbs} to the block Gibbs rejuvenation and block SMC setting, with user-specified blocking hints. These proposals are \textit{locally optimal} for models that contain only discrete finite-domain variables, meaning that of all possible proposals $Q_i$ they minimize the divergence 
$$KL(\pi_{i-1}(\mathbf{R}^{(i-1)}) Q_i(\Delta_i^\mathbf{R}; \mathbf{R}^{(i-1)}, d_i) || \pi_i(\mathbf{R}^{(i-1)} \cup \Delta_i^\mathbf{R})).$$
The distribution on the left represents a perfect sample $\mathbf{R}^{(i-1)}$ from the target given the first $i - 1$ observations, extended with the proposal $Q_i$. The distribution on the right is the target given the first $i$ data points.
In our setting the locally optimal proposal is given by
\begin{align*}
Q_i(\Delta_i^\mathbf{R};& \mathbf{R}^{(i-1)}, d_i) \propto \\
&p(\Delta_i^\mathbf{R} \mid \Delta_1^\mathbf{R}, \dots, \Delta_{i-1}^\mathbf{R})p(d_i \mid \Delta_1^\mathbf{R}, \dots, \Delta_{i}^\mathbf{R}).
\end{align*}
Algorithm~\ref{alg:increment-bayes-net} shows how to compile this distribution to a Bayesian network; when the latent attributes have finite domains, the normalizing constant can be computed and the locally optimal proposal can be simulated (and evaluated) exactly. 
This is possible because there are only a finite number of instantiations of the random increment $\Delta_i^\mathbf{R}$ to consider.
The compiler generates efficient enumeration code separately for each pattern of missing values it encounters in the dataset, exploiting conditional independence relationships in each Bayes net to yield potentially exponential savings over naive enumeration.
A similar strategy can be used to compile data-driven object-wise rejuvenation proposals, and to handle some continuous variables with conjugate priors; see supplement for details.

\subsection{Scaling to Large Models and Data with Inference Hints}
\label{sec:hints}
Scaling to models with large-domain variables and to datasets with many rows is a key challenge.
In PClean, users can specify lightweight \textit{inference hints} to the proposal compiler, shown in gray in Figure~\ref{fig:example-program}, to speed up inference without changing model's meaning.

\textbf{Programmable Subproblems.} First, users may group attribute and reference statements into blocks by wrapping them in the syntax  $\textbf{subproblem begin}\dots\textbf{end}$. This partitions the attributes and reference slots of a class into an ordered list of \textit{subproblems}, which SMC uses as intermediate target distributions. This makes enumerative proposals faster to compute, at the cost of considering less information at each step; rejuvenation moves can often compensate for short-sighted proposals.



{\bf Adaptive Mixture Proposals with Dynamic Preferred Values.}
A random variable within a model may be intractable to enumerate. For example, $\texttt{string\_prior(1, 100)}$ is a distribution over all strings between 1 and 100 letters long.
To handle these, PClean programs may declare \textit{preferred values hints}. Instead of $X \sim d(E,\dots,E)$, the user can write $X \sim d(E,\dots,E) \textbf{ preferring } E,$ where the final expression gives a list of values $\xi_X$ on which the posterior mass is expected to concentrate. 
%
When enumerating, PClean replaces the CPD $\phi_X$ with a surrogate $\hat{\phi}_X$, which is equal to $\phi_X$ for preferred value inputs in $\xi_X$, but 0 for all other values. The mass not captured by the preferred values, $1 - \sum_{x \in \xi_{X}} \phi_X(x)$, is assigned to a special $\textbf{other}$ token.
Enumeration yields a partial proposal $\hat{Q}$ over a modified domain; the full proposal $Q$ first draws from $\hat{Q}$ then replaces $\textbf{other}$ tokens with samples from the appropriate CPDs $\phi_X(\cdot \mid Pa(X))$. This yields a mixture proposal between the enumerative posterior on preferred values and the prior: when none of the preferred values explain the data well, $\textbf{other}$ will dominate, causing the attribute to be sampled from its prior. But if any of the preferred values are promising, they will almost certainly be proposed.

\begin{figure*}[t]
\includegraphics[width=0.95\linewidth]{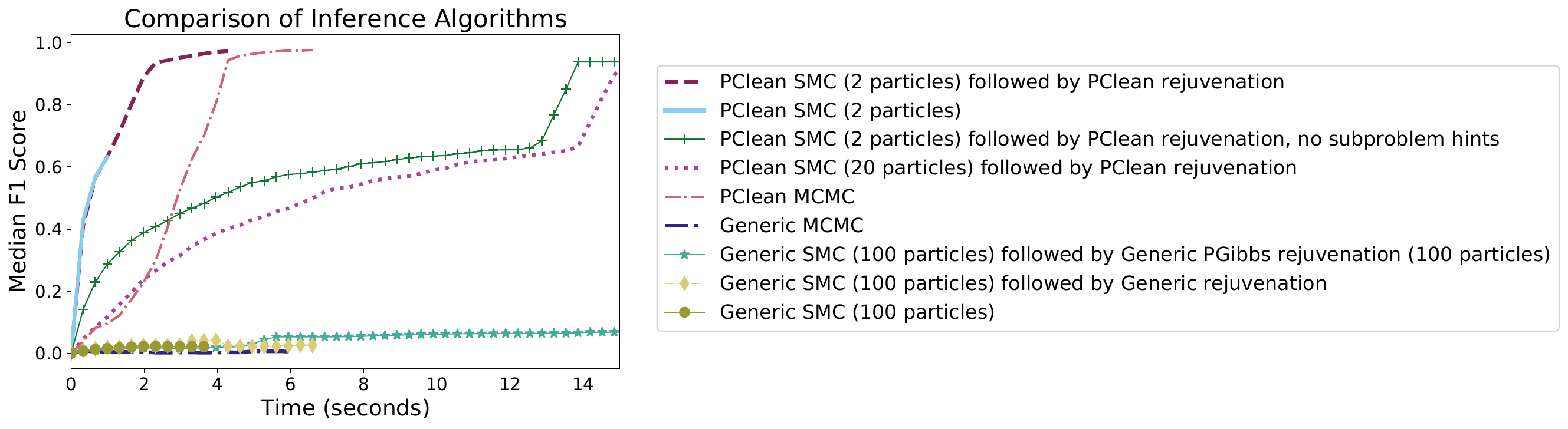}
\vspace{-4mm} 
\caption{Median accuracy vs. runtime for five runs of alternative inference algorithms on the \textit{Hospital} dataset \citep{Chu2013}, with an additional 20\% of cells artificially deleted so as to test both repair and imputation.}
\label{fig:inference-comparison}
\end{figure*}

\section{EXPERIMENTS}
\label{sec:results}
In this section, we demonstrate empirically that (1) PClean's inference works when standard PPL inference strategies fail, (2) short PClean programs suffice to compete with existing data cleaning systems in both runtime and accuracy, (3) PClean can scale to large real-world datasets, and (4) PClean's inference can deliver calibrated and useful estimates of uncertainty. In Experiments (1)-(3), we evaluate PClean's accuracy using a single posterior sample (the last iterate of PClean's final MCMC rejuvenation sweep); in Experiment (4), we consider an uncertainty-aware, multi-sample estimator of the clean dataset, which exploits our Bayesian framework for higher-precision repairs. Experiments were run on a  laptop with a 2.6 GHz CPU and 32 GB of RAM. 





\vspace{1.8mm}
\textbf{(1) Comparison to Generic PPL Inference.}
We evaluate PClean's inference against
standard PPL inference algorithms reimplemented
to work on PClean models, on a popular benchmark from the data cleaning literature (Figure~\ref{fig:inference-comparison}). We do not compare directly to other PPLs' implementations, because many (e.g. BLOG) cannot represent PClean's non-parametric prior. Some languages (e.g. Turing) have explicit support for non-parametric distributions, but could not express PClean's recursive use of CRPs. Others could in principle express PClean's model, but would complicate an algorithm comparison in other ways: Venture's dynamic dependency tracking is thousands of times slower than SOTA; Pyro's focus is on variational inference, hard to apply in PClean models; and Gen supports non-parametrics only via the use of mutation in its slower dynamic modeling language (making SMC $O(N^2)$) or via low-level extensions that would amount to reimplementing PClean using Gen's abstractions. Nonetheless, the algorithms in Figure~\ref{fig:inference-comparison} are inspired by the generic automated inference provided in many PPLs, which use top-down proposals from the prior for SMC, MH~\citep{dippl,ritchie2016c3}, and PGibbs~\citep{wood2014new,Murray2015,Mansinghka2014}. Our results show that PClean suffices for fast, accurate inference where generic techniques fail, and also demonstrate why inference hints are necessary for scalability: without subproblem hints, PClean takes much longer to converge, even though it eventually arrives at a similar $F_1$ value.

\begin{table*}[]
\centering
\begin{tabular}{|lc|ccccc|}
\hline
\multicolumn{1}{|l}{\textbf{\footnotesize{Task}}}    & \multicolumn{1}{l|}{\footnotesize{\textbf{Metric}}} & \multicolumn{1}{l}{\textbf{\footnotesize{PClean}}} & \multicolumn{1}{l}{\begin{tabular}[c]{@{}c@{}}\footnotesize{\textbf{HoloClean}}\\ \footnotesize{\textbf{(Unpublished)}} \end{tabular}}
 &
\multicolumn{1}{l}{\footnotesize{\textbf{HoloClean}
}}  &\multicolumn{1}{l}{\footnotesize{\textbf{NADEEF}
}} & \multicolumn{1}{l|}{\begin{tabular}[c]{@{}c@{}}\footnotesize{\textbf{NADEEF + Manual}}\\ \footnotesize{\textbf{Java Heuristics}} \end{tabular}} \\ \hline 
\multirow{2}{*}{\textbf{\footnotesize{Flights}}}        
& $F_1$   & \textbf{0.90}   & 0.64 & 0.41 & 0.07 & \textbf{0.90}
\\ 
 & {Time }   & \textbf{3.1s}  & { 45.4s}  & { 32.6s}  & { 9.1s} & 14.5s
 \\\hline
\multirow{2}{*}{\textbf{\footnotesize{Hospital}}}       
& $F_1$    & \textbf{0.91}  & 0.90  & 0.83 & 0.84 & 0.84  
\\
 & { Time }   & \textbf{{ 4.5s}}  & {1m 10s}  & { 1m 32s}  & { 27.6s} & 22.8s
 \\\hline
\multirow{2}{*}{\textbf{\footnotesize{Rents}}} 
& $F_1$    & \textbf{0.69}   &  0.48 & 0.48 & 0 & 0.51
\\ 
 & { Time }  & {1m 20s}  & { 20m 16s}  & {13m 43s}  & { 13s} & \textbf{{7.2s}}
 \\\hline
\end{tabular}
\caption{Results of PClean and various baseline systems on three diverse cleaning tasks.}
\vspace{-1mm}
\label{tab:results}
\end{table*}

\vspace{1.8mm}

\textbf{(2) Applicability to Data Cleaning.} 
To check PClean's modeling and inference capabilities are good for data cleaning \textit{in absolute terms} (rather than relative to generic PPL inference), we contextualize PClean's accuracy and runtime against two SOTA data-cleaning systems on three benchmarks with known ground truth (Table~\ref{tab:results}), described in detail in the supplement. Briefly, the datasets are \textit{Hospital}, a standard benchmark with artificial typos in 5\% of cells that can be corrected by leveraging duplication of entities across rows; \textit{Flights}, a standard benchmark integrating flight information (e.g. arrival, departure times) from conflicting real-world data sources; and \textit{Rents}, a synthetic dataset based on census statistics, featuring continuous and discrete values, misspelled county names, missing apartment sizes, and unit errors. The baseline systems are \textit{HoloClean}~\citep{Rekatsinas2017}, based on probabilistic machine learning, and \textit{NADEEF}, which uses a MAX-SAT solver to minimize violations of user-defined cleaning rules~\citep{Dallachiesat2013}. For HoloClean, we consider both the original code and the authors' latest (unpublished) version on GitHub; for NADEEF, we include results both using NADEEF's streamlined rule-definition interface and with custom, handwritten Java rules (more expressive but also more cumbersome).


Table~\ref{tab:results} reports \textit{$F_1$} scores and cleaning speed (see supplement for precision/recall). We do not aim to anoint a single `best cleaning system,’ since optimality depends on the available domain knowledge and the user's desired level of customization. Further, while we followed system authors’ per-dataset recommendations where possible, a pure system comparison is difficult, since each system relies on its own rule configuration. Rather, we note that short (\textless 50-line) PClean programs can encode knowledge useful in practice for cleaning diverse data, and inference is good enough to achieve $F_1$ scores as good or better than SOTA data-cleaning systems on all three datasets, often in less wall-clock time. As an illustration of the value and convenience of encoding relevant knowledge, on \textit{Flights}, a baseline, 16-line PClean program earns an $F_1$ score of $0.60$, but the $F_1$ can be boosted to 0.69 by encoding that sources have varying reliability (+1 line), and to 0.90 by encoding that for a given flight, an airline’s own website is most likely to be reliable (+1 line). By contrast, adding a similar reliability heuristic to NADEEF required 50 lines of Java; see supplement.

\textbf{(3) Scalability to Large, Real-World Data.}
We ran PClean on the Medicare Physician Compare National dataset, shown earlier in Figure~\ref{fig:physicians_results}.  It contains 2.2 million records, each listing a clinician and a practice location; the same clinician may work at multiple practices, and many clinicians may work at the same practice. NULLs and systematic errors are common (e.g. consistently misspelled city names at a practice).

PClean took 7h36m, performing 8,245 repairs and  1,535,415 imputations. Out of 100 randomly chosen imputations, 90\% agreed with manually obtained ground truth. We also verified that 7,954 repairs (96.5\%) were correct (some were correct normalization, e.g. choosing a single spelling for cities whose names could be spelled multiple ways). By contrast, NADEEF changed 88 cells across the whole dataset, and HoloClean did not initialize in 24 hours, using the configuration provided by HoloClean's authors.

Figure~\ref{fig:physicians_results} shows PClean's real behavior on four rows. Consider the misspelling \textit{Abington, MD}, which appears in 152 entries. The correct spelling \textit{Abingdon, MD} occurs in only 42. PClean still recognizes \textit{Abington} as an error, because all 152 instances share a single practice address, and errors are modeled as systematic at the practice level. Next, consider PClean's correct inference that Ryan's degree is \textit{DO}: more \textit{Family Medicine} doctors are MDs than DOs, but the school \textit{PCOM} awards many more DOs than MDs. All parameters enabling this reasoning are learned from the dirty data.

\begin{figure}
    \centering
    \includegraphics[width=\linewidth]{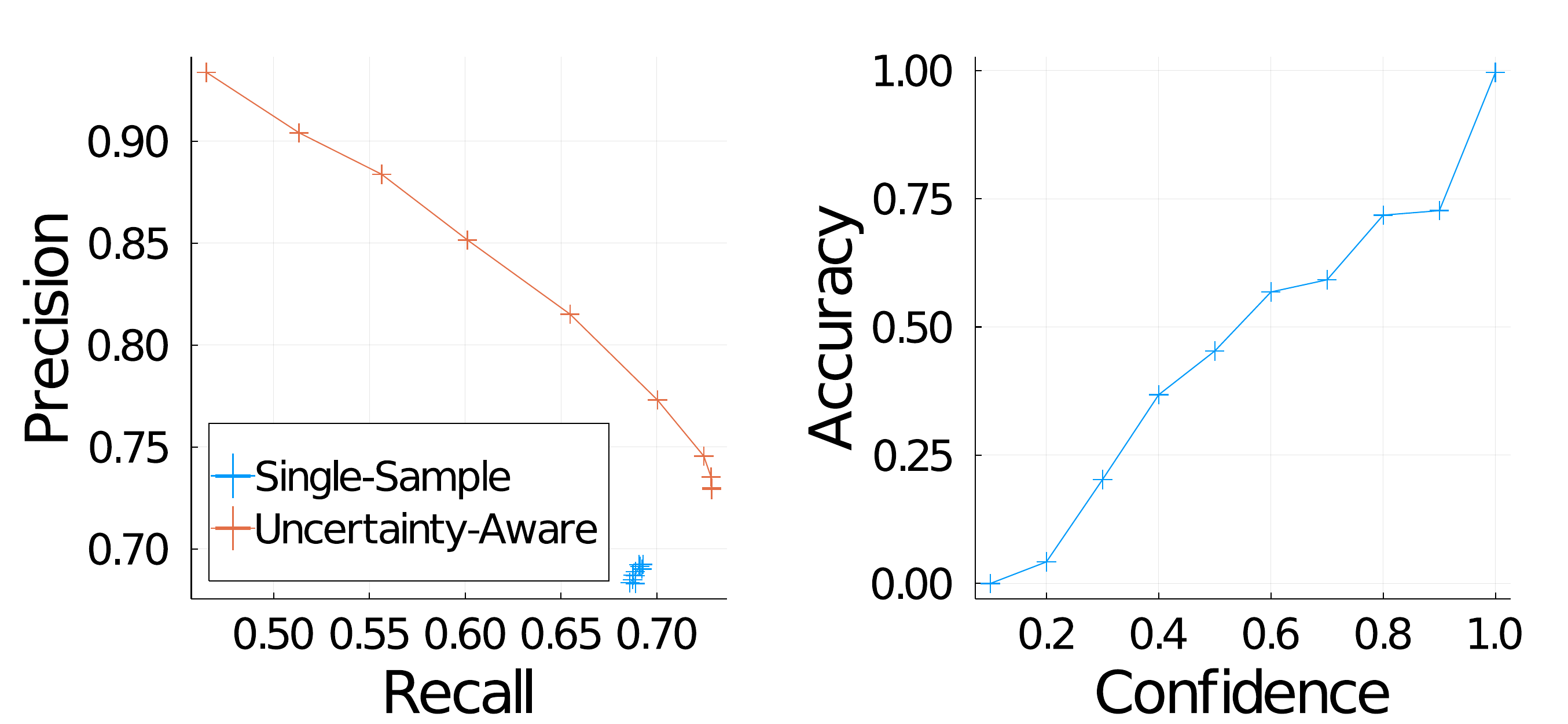}
    \caption{Exploiting Bayesian uncertainty with PClean on \textit{Rents}. Left: Ten independent posterior samples were generated. Blue marks show precision and recall achieved by considering each sample separately, whereas red curve shows precision vs. recall tradeoff when using modal predictions across samples, making repairs only when predictions surpass a confidence threshold. Right: PClean's uncertainty quantification appears to be well-calibrated on \textit{Rents}. `Confidence' is the proportion of independent posterior samples that agree on the modal predicted value for a cell; `accuracy' is the proportion of cells at a certain confidence level for which the modal prediction is correct.}
    \label{fig:calibration}
    \vspace{-4mm}
\end{figure}

\textbf{(4) MAP Estimation and Bayesian Uncertainty Quantification.} Our previous experiments used a single posterior sample to estimate the clean dataset. This experiment investigates strategies for exploiting richer information about the posterior, namely: (1) using the most common  predictions across multiple independent posterior samples (approximating the MAP clean dataset), and (2) setting a confidence threshold for repairs, to trade recall for higher precision. In particular, we ran PClean's inference with 10 parallel chains on the \textit{Rents} dataset, and collected 1 posterior sample from each (the last iterate of MCMC rejuvenation). Figure~\ref{fig:calibration}'s left panel shows, in blue, the precision and recall achieved by considering each sample individually, and in red, the various precision/recall tradeoffs achievable by using \textit{most common} prediction (across all 10 samples) for each cell, or leaving a cell unmodified if the confidence (i.e., the proportion of samples in which the modal value was predicted) does not surpass a threshold. The optimal $F_1$ of 0.73 ($R = 0.70, P = 0.77$), a 4-point improvement over the results from Table~\ref{tab:results}, is achieved by thresholding at $0.5$.
The right panel of Figure~\ref{fig:calibration} shows a calibration plot, obtained by binning the cells of the \textit{Rents} dataset into confidence levels and measuring the proportion of cells in each bin for which the most common prediction was correct. A caveat of our approach to uncertainty-aware cleaning is that performing independent repairs \textit{per cell} will not necessarily approximate the overall MAP clean dataset, unless the posterior is actually independent. Moreover, \textit{Rents} is a challenging but synthetic dataset; the degree to which these calibration results replicate in real-world problems will depend on the fidelity of the user's PClean program. However, our results suggest that PClean's inference engine is capable of delivering not only accurate cleaning results but also useful estimates of uncertainty.

\section{DISCUSSION}
PClean, like other domain-specific PPLs, aims to be more automated and scalable than general purpose PPLs, by leveraging structure in its restricted model class to deliver fast inference. At the same time, it aims to be expressive enough to concisely solve a broad class of real-world data cleaning problems.

Future development of PClean could build a more extensive standard library of primitives for modeling diverse data types (perhaps including neural models for text or image data), and a more robust proposal compiler for free-text and continuous latent variables (perhaps based on learning neural proposals for selected attributes). Integrating PClean with the abstractions of a mature probabilistic programming system, such as Gen, could facilitate implementation. PClean's scalability could also be improved, by exploring distributed variants of PClean based on Bayesian formulations of blocking~\citep{marchant2021d}. A more speculative research direction is to (partially) automate PClean program authoring, by applying techniques such as automated error modeling \citep{Heidari2019} or probabilistic program synthesis~\citep{saad-popl-2019,choi2020group}. It could also be fruitful to develop hierarchical variants of PClean that enable learned parameters and latent objects to transfer across datasets.

PClean could be described as a {\it data-driven} probabilistic expert system~\citep{horvitz1988decision,pearl1988probabilistic,heckerman1992toward,shafer1996probabilistic}, incorporating ideas from probabilistic programming to scale to messy, real-world domain knowledge and data.
Crucially, since PClean can infer the objects and parameters of a domain from data, users need only encode higher-level domain knowledge, not brittle details. 
It remains to be seen whether systems like PClean can be made to give meaningful explanations of individual judgments (like those offered by human experts).

Our results show that probabilistic programs can clean dirty, denormalized data with state-of-the-art accuracy and performance. More broadly, PClean joins existing domain-specific PPLs in demonstrating that it is feasible and useful to integrate sophisticated styles of modeling and inference, developed over years of research, into simple languages and specialized inference engines. We hope PClean proves useful to practitioners, and that it encourages researchers to develop new domain-specific PPLs for other important problems.

\subsubsection*{Acknowledgements}
The authors are grateful to Zia Abedjan, Marco Cusumano-Towner, Raul Castro Fernandez, Cameron Freer, Divya Gopinath, Christina Ji, Tim Kraska, George Matheos, Feras Saad, Michael Stonebraker, Josh Tenenbaum, and Veronica Weiner for useful conversations and feedback, as well as to our anonymous referees for their constructive suggestions. This work is supported by the National Science Foundation Graduate Research Fellowship Program under Grant No. 1745302; DARPA, under the Machine Common Sense (MCS) and Synergistic Discovery and Design (SD2) programs; gifts from the Aphorism Foundation and the Siegel Family Foundation; a research contract with Takeda Pharmaceuticals; and financial support from Facebook, Google, and the Intel Probabilistic Computing Center.



\bibliography{pclean}

\onecolumn

\algblockdefx[model]{Model}{EndModel}%
[2]{\textsc{#1}(#2):}{}

\addvspace{-5cm}
\appendix


\section{Baseline Inference Algorithms}
\label{sec:baseline-algs}

The paper's Figure 6 shows median accuracy vs. time for five independent runs of nine inference algorithms. These results were computed using the PClean program shown in Appendix~\ref{sec:hospital-pclean-program}, on a version of the \textit{Hospital} dataset (Appendix~\ref{sec:benchmarks}) with 20\% of its cells deleted at random, to test both repair and imputation (the original \textit{Hospital} dataset has many errors, but very few missing cells). Below, we give descriptions of each inference algorithm we test:

\begin{enumerate}
    \item {\bf PClean SMC (2 particles) followed by PClean rejuvenation} is the inference algorithm described in Section 3. First, a complete run of 2-particle sequential Monte Carlo, using PClean's enumeration-based compiled proposals, is completed, incorporating all 1000 rows of the dataset. Then, one of the two particles is selected, and for each object in its latent database, a block rejuvenation MCMC kernel is run, also using PClean's enumeration-based compiled proposal. (The number of MCMC moves completed during this sweep will depend on the number of objects inferred for the latent database, a quantity that varies from run to run. See note below this list for an explanation of how median accuracies were computed across runs with different numbers of iterations.)
    \item {\bf PClean SMC (2 particles)} is the same as the above except that no rejuvenation sweep is performed.
    \item {\bf PClean SMC (2 particles) followed by PClean rejuvenation, no subproblem hints} is the same as (1), except we disregard subproblem hints in the PClean program. (The program in question, shown in Appendix~\ref{sec:hospital-pclean-program}, has two subproblem hints.) As a result, SMC takes bigger steps, and enumerative proposals take longer to execute (but are higher quality).
    \item {\bf PClean SMC (20 particles) followed by PClean rejuvenation} is the same as (1) except with 20 particles, instead of 2.
    \item {\bf PClean MCMC} initializes the latent database using ancestral sampling, i.e., from the prior, but modified to use observed values when they are available. It then performs two complete MCMC sweeps, using PClean's block rejuvenation proposals; each sweep performs an MCMC move for each object in the current latent database.
    \item {\bf Generic MCMC} initializes the latent database as in (5), and performs ten complete sweeps using \textit{single-site} Metropolis-Hastings (tens of thousands of accept/reject steps). That is, each individual attribute or reference slot is separately updated, using the prior as proposal. When a reference slot is proposed, there is a chance that a new object is also proposed as its target. We note that our implementation is much faster than most PPLs' single-site Metropolis-Hastings implementations, as it re-evaluates only those likelihood terms affected by the proposed single-variable change.
    
    \item {\bf Generic SMC (100 particles) followed by generic PGibbs rejuvenation (100 particles)} initializes the latent database using 100-particle sequential Monte Carlo, using the same sequence of target distributions as in PClean SMC, but with the prior as a proposal. This is followed by three sweeps of Particle Gibbs rejuvenation moves: as in PClean rejuvenation from (1), we perform per-\textit{object} updates, but the proposal is generated not via PClean's enumerative proposal compiler, but rather by using 100-particle conditional sequential Monte Carlo (CSMC) [like a Gibbs move, this proposal is always accepted]. We note that this baseline improves over existing PPLs' support for Particle Gibbs in several ways. First, Particle Gibbs updates only those variables connected to a particular latent object, rather than trying to update the entire model state at once. Second, incremental SMC weights are computed incrementally, evaluating only those likelihood terms that are necessary. Third, a reweighting (and, based on ESS, possibly resampling) step is triggered whenever a new likelihood term could possibly be evaluated, regardless of how the PClean program is written. However, unlike PClean's rejuvenation moves (but like many other PPL implementations), our ``generic PGibbs rejuvenation'' uses proposals from the prior for its CSMC sweeps, greatly limiting its effectiveness. (We note that \textit{delayed sampling}~\cite{murray2018delayed,wigren2019parameter} is a sophisticated PPL technique that could provide benefits similar to those provided by PClean's proposal; however, to our knowledge delayed sampling is not implemented in any PPL capable of performing SMC in PClean's model.)
    
    \item {\bf Generic SMC (100 particles) followed by generic rejuvenation} initializes the latent database using 100-particle sequential Monte Carlo, as in (7). It then performs five single-site Metropolis-Hastings rejuvenation sweeps (tens of thousands of accept/reject steps), as described in (6).
    
    \item {\bf Generic SMC (100 particles)} initializes the latent database as in (7) and performs no additional rejuvenation.
\end{enumerate}

For each run of each algorithm, time and accuracy were measured after each SMC step or MCMC transition. Since steps/transitions finished at different timestamps across runs, and because each run of an algorithm lasted a different number of steps (due to the stochastic number of objects in the latent database), we used linear interpolation to approximate a continuous time/accuracy curve for each run. Then, to plot median performance across the five runs, we took the median value across the interpolated curves at a fixed set of times. In all nine algorithms, all five runs ended at roughly the same time; the plotted endpoint for each algorithm was chosen as the time when the {\it last} run was complete. For any run that finished slightly earlier, the accuracy value was extrapolated as the accuracy at its last timestamp.

\section{Evaluation on Data Cleaning Benchmarks: Datasets, Systems, and System Configurations}

Table 1 of our paper provides evidence of PClean's applicability to data-cleaning problems, by comparing accuracy and runtime for three PClean programs against state-of-the-art data cleaning systems applied to the same benchmark datasets. The table reports $F_1$ scores, but omits the breakdown in terms of recall ($R=\frac{\text{correct repairs}}{\text{total errors}}$) and precision ($P = \frac{\text{correct repairs}}{\text{total repairs}}$), the metrics from which $F_1 = 2PR/(P + R)$ is derived. The table below presents a fuller picture:

\begin{center}
\begin{tabular}{|lc|ccccc|}
\hline
\multicolumn{1}{|l}{\textbf{\footnotesize{Task}}}    & \multicolumn{1}{l|}{\footnotesize{\textbf{Metric}}} & \multicolumn{1}{l}{\textbf{\footnotesize{PClean}}} & \multicolumn{1}{l}{\begin{tabular}[c]{@{}c@{}}\footnotesize{\textbf{HoloClean}}\\ \footnotesize{\textbf{(Unpublished)}} \end{tabular}}
 &
\multicolumn{1}{l}{\footnotesize{\textbf{HoloClean}
}}  &\multicolumn{1}{l}{\footnotesize{\textbf{NADEEF}
}} & \multicolumn{1}{l|}{\begin{tabular}[c]{@{}c@{}}\footnotesize{\textbf{NADEEF + Manual}}\\ \footnotesize{\textbf{Java Heuristics}} \end{tabular}} \\ \hline 
\multirow{2}{*}{\textbf{\footnotesize{Flights}}}        
& Prec  & {0.91}   & 0.79  & 0.39 & 0.76  & 0.92 \\ 
& Rec   & {0.89}   & 0.55 & 0.45 & 0.03  & 0.88  \\ 
& $F_1$   & \textbf{0.90}   & 0.64 & 0.41 & 0.07 & \textbf{0.90}
\\ 
 & {Time }   & \textbf{3.1s}  & { 45.4s}  & { 32.6s}  & { 9.1s} & 14.5s
 \\\hline
\multirow{2}{*}{\textbf{\footnotesize{Hospital}}}       
& Prec  & {1.0}   & 0.95  & {1.0} & 0.99   & 0.99 \\ 
& Rec  & 0.83 & {0.85}  & 0.71 & 0.73  & 0.73 \\ 
& $F_1$    & \textbf{0.91}  & 0.90  & 0.83 & 0.84 & 0.84  
\\
 & { Time }   & \textbf{{ 4.5s}}  & {1m 10s}  & { 1m 32s}  & { 27.6s} & 22.8s
 \\\hline
\multirow{2}{*}{\textbf{\footnotesize{Rents}}} 
& Prec    & 0.68  & {0.83} &  {0.83} & 0  & {0.83}   \\ 
& Rec    & {0.69}  &  0.34  &  0.34 &  0 & 0.37 \\ 
& $F_1$    & \textbf{0.69}   &  0.48 & 0.48 & 0 & 0.51
\\ 
 & { Time }  & {1m 20s}  & { 20m 16s}  & {13m 43s}  & { 13s} & \textbf{{7.2s}}
 \\\hline
\end{tabular}
\end{center}

The remainder of this appendix describes in detail: each benchmark dataset (Appendix~\ref{sec:benchmarks}), each baseline system (Appendix~\ref{sec:baseline-systems}), the HoloClean and NADEEF configurations used for each baseline (emphasizing the ways in which we attempted to encode dataset-specific domain knowledge) (Appendix~\ref{sec:configurations}), and the PClean programs we used for each dataset (Appendix~\ref{sec:programs}).

\subsection{Description of Benchmarks}
\label{sec:benchmarks}
The three smaller benchmark datasets are available in the PClean repository; \textit{Physicians} is excluded for size, but is hosted by Medicare.

\textit{Hospital} is a real-world Medicare dataset, but with artificially introduced typos in approximately 5\% of its 19,000 cells (1000 rows, 19 columns). Each row reports the performance of a particular hospital on a particular metric, and it includes metadata such as hospital address and phone number. This leads to a lot of duplicated information, as the same hospital appears multiple times (with different metrics), and the same metrics also appear multiple times (with different hospitals). All this duplication facilitates accurate cleaning even in the presence of typos. 

\textit{Flights} consists of 2,377 rows describing real-world flight, their scheduled departure/arrival times, and their true departure/arrival times, as scraped from the web. These times often conflict between the sources, so the task is to integrate them to form a consistent dataset.  We use the version from \cite{Mahdavi2019}.

\textit{Rents} is a new synthetic dataset of apartment listings that we derived from census and housing statistics \cite{USCensusBureau2019}. It contains bedroom size, rent, county, and state.  We first generated a clean dataset with 50,000 rows in the following manner:
\begin{itemize}
    \item The county-state combination is chosen proportionally to its population in the United State
    \item The size of the apartment is chosen uniformly from studio, 1 bedroom, 2 bedroom, 3 bedroom, 4 bedroom.
    \item The rent is chosen according to a normal distribution in which the mean is the median rent for an apartment of the chosen size in the chosen country and the standard deviation is chosen to be 10\% of the mean
\end{itemize}
The dataset was then dirtied in the following ways: 
\begin{itemize}
    \item 10\% of state names are deleted (many counties exist across multiple states, e.g. 30 states have a Washington County).
    \item Approximately 1-2\% of county names are misspelled
    \item 10\% of apartment sizes are deleted
    \item 1\% of apartment prices are listed in the incorrect units (thousands of dollars, instead of dollars)
\end{itemize}


\subsection{Description of State-of-the-Art Data-Cleaning Systems}
\label{sec:baseline-systems}

\textit{HoloClean} is a data-cleaning system, which compiles user-provided \textit{integrity constraints} and when available, external ground-truth, into a factor graph with learned weights ~\cite{Rekatsinas2017}. These integrity constraints describe cells that should match, conditional on the agreement of other fields, e.g. if zip codes of two rows match, the states in those two rows should match. These constraints can also be made with respect to external data (e.g. if a row's zip code in the table matches a zip code in a gazetteer, the row's state should match the corresponding state in the gazetteer). 

\textit{NADEEF} is a data-cleaning system that leverages user-specified cleaning rules ~\cite{Dallachiesat2013}. NADEEF compiles users' rules into a weighted MAX-SAT query and runs it through a solver, then uses the results to clean the data. User-specified rules can either be integrity constraints (as HoloClean) or handcrafted rules. These handcrafted rules take the form of Java classes, in which users write a \texttt{detect} function that takes in a pair of tuples and outputs whether one or more violations have been detected, and if so, over which groupings of cells. The user can also optionally write a \texttt{repair} function that takes in those detected cells, and returns a fix. That is, unlike in PClean, user-encoded knowledge explicitly describes how to both detect and repair violations.

To our knowledge, neither system comes with special logic for handling text fields, dates, etc. as distinct from general categorical data. 

\subsection{Settings for Data-Cleaning Systems}
\label{sec:configurations}
Below, we present the integrity constraints we encoded in both HoloClean and NADEEF, as well as the handcrafted Java rules for NADEEF. The integrity constraints are presented as $``A \textnormal{ determines } B"$, which means that for two rows, if all columns in $A$ match, one should expect all columns in $B$ to also match.

For each NADEEF Java rule, we describe the functionality and report the number of lines of code used to encode it (ignoring imports, boilerplate, and parentheses). All integrity constraints and Java rules can also be found in the PClean repository.

\textbf{On encoding domain knowledge.} Data cleaning is of course easier with accurate domain knowledge about the data and the likely errors. This is one reason we developed PClean: to enable generatively encoded domain knowledge to inform a data cleaning system. This does, however, raise the question of how to compare PClean fairly to other data-cleaning systems: if PClean is more accurate only because it encodes more domain knowledge, it would be misleading to claim that PClean is `better' in some absolute sense than an existing system. Our evaluation in Section 4 specifically explains that this is not our intention: we just mean to contextualize PClean's accuracy and runtime in the context of other data-cleaning systems, using reasonable configurations for those systems.

That said, we tried our best to encode as much helpful domain knowledge as we could into the configurations for HoloClean and NADEEF. Some of the settings below were chosen in response to direct advice from authors of each system; others were based on existing scripts, written by the system authors, for cleaning these benchmark datasets (some of our benchmarks also appeared in the papers presenting these systems). In addition, we tried tweaking these configurations ourselves, and reported the best numbers we could. 

It is likely that the \textit{approaches} that NADEEF and HoloClean take, of using weighted logic and factor graphs, could in principle express richer domain knowledge than our configurations here encode. But to our knowledge, the current \textit{systems} do not expose these capabilities in easy-to-exploit ways. 


\subsubsection{Hospital}
\textbf{Integrity Constraints}
\begin{itemize}
    \item \textit{Hospital Name} determines \textit{Phone Number}, \textit{City}, \textit{ZIP Code}, \textit{State}, \textit{Address}, \textit{Provider Number}, \textit{County Name}, \textit{Hospital Type},  and \textit{Hospital Owner}.
    \item \textit{Phone Number} determines \textit{City}, \textit{ZIP Code}, \textit{State}, \textit{Address1}, \textit{Provider Number}, \textit{County Name}, \textit{Hospital Type}, \textit{Hospital Owner}.
    \item \textit{ZIP Code} determines \textit{City} and  \textit{State}.
    \item \textit{Measure Code} determines \textit{Measure Name} and \textit{Condition}.
    \item \textit{Measure Code} and \textit{State} together determine \textit{State Average}.
    
\end{itemize}

\textbf{Java Rules}

 The \textit{State Average} field is a concatenation of the \textit{Measure Code} and \textit{State} fields. For any row, we raise a violation if the concatenation does not hold over those three cells. We do not provide a repair, since it's unclear from that row alone which of the three cells is the incorrect one. This took 9 lines of Java code.

\subsubsection{Flights}

\textbf{Integrity Constraints}
\begin{itemize}
    \item \textit{Flight} number determines both the \textit{Scheduled Departure Time} and the \textit{Actual Departure Time}
    \item \textit{Flight} number determines both the \textit{Scheduled Departure Time} and the \textit{Actual Departure Time}
    
\end{itemize}

\textbf{Java Rules}

For a pair of rows, if both flights have the same flight number, a violation is already raised by the existing integrity constraints if the departure or arrival time does not match. The source corresponding to the flight's airline tends to more correct than third-party sources. Therefore, when applicable over a pair of rows, we provided the suggested repair of choosing the time from the website of the airline. This took 52 lines of Java code.

\subsubsection{Rent}
\textbf{Integrity Constraints}

\textit{County} determines \textit{State}. 

\textbf{Java Rules}

If a state was missing for a rental listing, we suggested that NADEEF choose the repair of the most common state corresponding to a given county (which it would not otherwise do), requiring 48 lines of Java.

Additionally, if a rent was below a certain fixed threshold, the program would flag as a violation, and multiply by the correct factor for a unit conversion. This second rule required 12 lines of Java. 

\subsubsection{Physician}
\textbf{Integrity Constraints}
\begin{itemize}
\item The \textit{National Provider Identifier (NPI)} determines the \textit{PAC ID} and vice versa.
\item The \textit{National Provider Identifier (NPI)} determines \textit{First Name}, \textit{Last Name}, \textit{Medical School Name}, and \textit{Graduation Year}.
\item The \textit{Group Practice ID} determines the \textit{Organization name}.
\item The \textit{Zip Code} determines the \textit{City} and \textit{State}.
\end{itemize}

\subsection{PClean Programs}
\label{sec:programs}

In this section, we present the PClean programs we used to clean each benchmark dataset. This is the closest analogue to a `configuration' of an automated data-cleaning system. But rather than encode rules for detecting and repairing errors, PClean programs encode generative models of relational databases and of the process by which they are corrupted, filtered, and joined to yield flat, dirty, denormalized datasets.

\subsubsection{Hospital}
\label{sec:hospital-pclean-program}
The \textit{Hospital} dataset is modeled with seven classes: \texttt{Record}s reflect typo'd attributes of \texttt{Hospital}s and the \texttt{Measure}s by which they are evaluated; \texttt{Hospital}s have \texttt{HospitalType}s and are located in \texttt{Place}s; \texttt{Place}s belong to \texttt{County} objects; and each \texttt{Measure} is related to some \texttt{Condition}. Typos are modeled as independently introduced for each cell of the dataset. Some fields are modeled as draws from broad priors over strings, whereas others are modeled as categorical draws whose domain is the set of unique observed values in the relevant column (some of which are in fact typos).

Inference hints are used to focus proposals for \texttt{string\_prior} choices 
on the set of strings that have actually been observed in a given column, and 
also to set  a custom subproblem decomposition for the \texttt{Record} class (all other classes use the default decomposition). 

\begin{lstlisting}[style=PClean,numbers=none]
latent class County
  parameter state_proportions ~ dirichlet(ones(num_states))
  state ~ discrete(observed_values[:State], state_proportions)
  county ~ string_prior(3, 30) preferring observed_values[:CountyName]
end
latent class Place
  county ~ County
  city ~ string_prior(3, 30) preferring observed_values[:City]
end
latent class Condition
  desc ~ string_prior(5, 35) preferring observed_values[:Condition]
end
latent class Measure
  code ~ uniform(observed_values[:MeasureCode])
  name ~ uniform(observed_values[:MeasureName])
  condition ~ Condition
end
latent class HospitalType
  desc ~ string_prior(10, 30) preferring observed_values[:HospitalType]
end
latent class Hospital
  parameter owner_dist ~ dirichlet(ones(num_owners))
  parameter service_dist ~ dirichlet(ones(num_services))
  loc ~ Place
  type ~ HospitalType
  id ~ uniform(observed_values[:ProviderNumber])
  name ~ string_prior(3, 50) preferring observed_values[:HospitalName]
  addr ~ string_prior(10, 30) preferring observed_values[:Address1]
  phone ~ string_prior(10, 10) preferring observed_values[:PhoneNumber]
  owner ~ discrete(observed_values[:HospitalOwner], owner_dist)
  zip ~ uniform(observed_values[:ZipCode])
  service ~ discrete(observed_values[:EmergencyService], service_dist)
end
latent class Record
  subproblem begin
    hosp   ~ Hospital;                      service ~ typos(hosp.service)
    id     ~ typos(hosp.id);                name    ~ typos(hosp.name)
    addr   ~ typos(hosp.addr);              city    ~ typos(hosp.loc.city)
    state  ~ typos(hosp.loc.county.state);  zip     ~ typos(hosp.zip)
    county ~ typos(hosp.loc.county.county); phone   ~ typos(hosp.phone)
    type   ~ typos(hosp.type.desc);         owner   ~ typos(hosp.owner)
  end
  subproblem begin
    metric ~ Measure
    code ~ typos(metric.code); mname ~ typos(metric.name);
    condition ~ typos(metric.condition.desc)
    stateavg = "$(hosp.loc.county.state)_$(metric.code)"
    stateavg_obs ~ typos(stateavg)
  end
end
\end{lstlisting}

\subsubsection{Flights}

The model for \textit{Flights} uses three classes: each observed \texttt{Record} comes from a \texttt{TrackingWebsite} and is about a \texttt{Flight}:

\begin{lstlisting}[style=PClean,numbers=none]
latent class TrackingWebsite
  name ~ string_prior(2, 30) preferring observed_values[:website]
end
latent class Flight
  flight_id ~ string_prior(10, 20) preferring flight_ids; index on flight_id
  sdt ~ time_prior() preferring observed_values["$flight_id-sched_dep_time"]
  sat ~ time_prior() preferring observed_values["$flight_id-sched_arr_time"]
  adt ~ time_prior() preferring observed_values["$flight_id-act_dep_time"]
  aat ~ time_prior() preferring observed_values["$flight_id-act_arr_time"]
end
latent class Record
  parameter error_probs[_] ~ beta(10, 50)
  flight ~ Flight; src ~ TrackingWebsite
  error_prob = lowercase(src.name) == lowercase(flight.flight_id[1:2]) ? 1e-5 : error_probs[src.name]
  sdt ~ maybe_swap(flight.sdt, observed_values["$(flight.flight_id)-sched_dep_time"], error_prob)
  sat ~ maybe_swap(flight.sat, observed_values["$(flight.flight_id)-sched_arr_time"], error_prob)
  adt ~ maybe_swap(flight.adt, observed_values["$(flight.flight_id)-act_dep_time"],   error_prob)
  aat ~ maybe_swap(flight.aat, observed_values["$(flight.flight_id)-act_arr_time"],   error_prob)
end
\end{lstlisting}

In the \textbf{parameter} declaration for \textit{error\_probs}, we use the syntax \texttt{error\_probs[\_] $\sim$ beta(10, 50)} to introduce a \textit{collection} of parameters; the declared variable becomes a dictionary, and each time it is used with a new index, a new parameter is instantiated. We use this to learn a different \texttt{error\_prob} parameter for each tracking website. We could alternatively declare \texttt{error\_prob} as an attribute of the \texttt{TrackingWebsite} class. However, PClean's inference engine uses smarter proposals for declared \textbf{parameter}s (taking advantage of conjugacy relationships), so for our experiments, we use the \textbf{parameter} declaration instead. We hope to extend automatic conjugacy detection to all attributes, not just parameters, in the near future.

As in \textit{Hospital}, we use \texttt{observed\_values} to provide inference hints to the broad \texttt{time\_prior}; this expresses a belief that the true timestamp for a certain field is likely one of the timestamps that has actually been observed, in the dirty dataset, with the given flight ID.

\subsubsection{Rents}
The program we use for \textit{Rents} contains two classes: \texttt{Listing}s are for apartments in some \texttt{County}:
\begin{lstlisting}[style=PClean, numbers=none]
data_table.block = map(x -> "$(x[1])$(x[end])", data_table.County)
units = [Transformation(identity, identity, x -> 1.0),
         Transformation(x -> x/1000.0, x -> x*1000.0, x -> 1/1000.0)]
latent class County
  parameter state_pops ~ dirichlet(ones(num_states))
  block ~ unmodeled(); index by block
  name ~ string_prior(10, 35) preferring observed_values[block]
  state ~ discrete(states, state_pops)
end
latent class Listing
  parameter avg_rent[_] ~ normal(1500, 1000)
  subproblem begin
    county ~ County
    county_name ~ typos(county.name, 2)
    br ~ uniform(room_types)
    unit ~ uniform(units)
    rent_base = avg_rent["$(county.state)_$(county.name)_$(br)"]
    observed_rent ~ transformed_normal(rent_base, 150.0, unit)
  end
  rent = round(unit.backward(observed_rent))
end
\end{lstlisting}

We model the fact that the rent may be in \textit{grand} instead of \textit{dollars}, as well as that the county name may contain typos. 
We introduce an artificial field, \textit{block}, consisting of the
first and last letters of the observed (possibly erroneous) County
field, and use it to inform an inference hint: we hint that posterior mass 
for a county's name concentrates on those strings observed somewhere in the dataset that share a first
and last letter in common with the observed county name for this row. Without
this approximation, inference is much slower (but potentially more accurate).

\subsubsection{Physicians}

The model for \textit{Physicians} contains five classes: \texttt{Record}s reference \texttt{Practice}s and \texttt{Physician}s; each \texttt{Physician} attended some medical \texttt{School}; and each \texttt{Practice} is in a \texttt{City}:

\begin{lstlisting}[style=PClean, numbers=none]
latent class School
  name ~ unmodeled(); index by name
end

latent class Physician
  parameter error_prob ~ beta(1.0, 1000.0)
  parameter degree_proportions[_] ~ dirichlet(3 * ones(num_degrees))
  parameter specialty_proportions[_] ~ dirichlet(3 * ones(num_specialties))
  npi ~ number_code_prior()
  school ~ School
  subproblem begin
    degree ~ discrete(observed_values[:Credential], degree_proportions[school.name])
    specialty ~ discrete(observed_values[Symbol("Primary specialty")], specialty_proportions[degree])
    degree_obs ~ maybe_swap(degree, observed_values[:Credential], error_prob)
  end
end

latent class City
  c2z3 ~ unmodeled(); index by c2z3
  name ~ string_prior(3, 30) preferring cities[c2z3]
end

latent class Practice
  addr ~ unmodeled(); index by addr
  addr2 ~ unmodeled(); index by addr2
  zip ~ string_prior(3, 10); index by zip
  legal_name ~ unmodeled(); index by legal_name
  subproblem begin
    city ~ City
    city_name ~ typos(city.name)
  end
end

latent class Record
  physician ~ Physician
  address ~ Practice
end
\end{lstlisting}

Many columns are not modeled. Similar to \textit{Rents}, we use a \textbf{parameter} in the \texttt{Physician} class for \texttt{degree\_probs}, although it might seem more natural to use an attribute of the \texttt{School} class; the resulting model is the same, but using \textbf{parameter} allows PClean to exploit conjugacy.

\subsection{Effect of Additional Domain Knowledge}
The quality of PClean's inference depends on the PClean program one uses to model the data. To demonstrate this, we apply four different PClean programs on \textit{Flights}. In our baseline (16 lines of code), we assume all sources are equally reliable and achieve an F1 score of 0.56. By additionally modeling the timestamp format, we achieve an F1 of 0.60. If we program PClean to learn a per-source reliability (one extra line of code), F1 climbs to 0.69. Finally, if we provide our program that the airline's own website is likely to be the most reliable for a given flight (one additional line of code for a total of 18), F1 jumps to 0.90.

We also implemented a user-defined cleaning rule in NADEEF, manually specifying a repair procedure for flight times that searched for a reported time from the flight's airline, and used that if available. This rule enabled NADEEF to clean the \textit{Flights} data, but required 52 lines of Java (beyond the boilerplate required for every NADEEF rule). Furthermore, as Table 1 of the paper shows, even encoding manual Java rules is, for some datasets, not enough to yield accurate cleaning.

\section{Additional Model Details}
\label{sec:modeling-details}
\subsection{Discrete Random Measure representation}

Our non-parametric structure prior $p(\mathbf{S})$ is described by Section 2 of the paper in terms of the two-parameter Chinese Restaurant Process. It is also possible to represent the generative process encoded by a PClean program using the Pitman-Yor process:
\\

\begin{algorithmic}[0]
\Model{GenerateDataset}{}
\For{latent class $C \in \textsc{TopologicalSort}(\mathcal{C})$}
  \State $\theta_C \sim p_{\theta_C}()$ 
  \State $G_C \sim \textsc{GenerateCollection}(C, \theta_C, \{G_{C'}\}_{C' \in Pa(C)})$
\EndFor
\State $\theta_\textbf{Obs} \sim p_{\theta_{\textbf{Obs}}}()$
\For{$i \in \{1, \dots, n\}$}
  \State $r_i \sim \textsc{GenerateObject}(C_{Obs}, \theta_{C_{Obs}}, \{G_C\}_{C \in Pa(C_{Obs})})$
\EndFor
\EndModel
\Model{GenerateCollection}{$C, \theta_C, \{G_{C'}\}_{C' \in Pa(C)}$}
\State $s_C \sim \textit{Gamma}(1, 1)$ 
\State $d_C \sim \textit{Beta}(1, 1)$ 
\State $G_C \sim PY(s_C, d_C, \textsc{GenerateObject}(C, \theta_C, \{G_{C'}\}_{C' \in Pa(C)}))$
\EndModel
\Model{GenerateObject}{$C,\theta_C, \{G_{C'}\}_{C' \in Pa(C)}$}
\For{reference slot $Y \in \mathcal{R}(C)$}
\State $r.Y \sim G_{T(C.Y)}$ 
\EndFor
\For{attribute $X \in \mathcal{A}(C)$}
\State $r.X \sim \phi_{C.X}(\theta_C, \{r.\tau\}_{\tau \in Pa(C.X)})$
\EndFor
\EndModel
\end{algorithmic}

We process classes one at a time, in topological order. For each latent class, we (1) generate
class-wide hyperparameters $\theta_C$ from their corresponding hyperpriors, and (2) generate an \textit{infinite weighted
collection} of \textit{objects} of class $C$. In this setting, an \textit{object} $r$ \textit{of class} $C$ is an assignment of
each attribute $C.X$ to a value $r.X$ and of each reference slot $C.Y$ to an object $r.Y$ of class $T(C.Y)$.
An infinite collection of latent objects is generated via a Pitman-Yor Process~\cite{Teh2011}:
$$G_C \sim PY(s_C, d_C, \textsc{GenerateObject}(C, \theta_C, \{G_{C'}\}_{C' \in Pa(C)}))$$
The Pitman-Yor Process is a discrete random measure that generalizes the Dirichlet Process. 
It can be understood as first sampling an infinite vector of probabilities 
$\rho \sim GEM(s_C, d_C)$ from a two-parameter GEM distribution, then setting $G_C = 
\sum_{i=1}^\infty \rho_i \delta_{r^C_i}$, where each of the infinitely many objects 
$r^C_i$ is distributed according to $\textsc{GenerateObject}(C, \theta_C, \{G_{C'}\}_{C' \in Pa(C)})$. This itself is a distribution over \textit{objects}, which first samples reference slots and then attributes.
   
To generate the objects of the observation class, which will be translated by the program's query into the flat dataset \textbf{D}, we sample $\theta_{C_{Obs}}$ from its prior distribution, then, for $i \in \{1, \dots, n\}$, generate the $i^{\text{th}}$ observed entry: $r_i \sim \textsc{GenerateObject}(C_{Obs}, \theta_{C_{Obs}}, \{G_C\}_{C \in Pa(C_{Obs})})$.

\subsection{Description of primitive distributions}

Our models for particular datasets make use of PClean's built-in probability distributions, which include not just the common distributions for categorical and numerical data, but also several domain-specific distributions useful for modeling strings and random errors. We briefly summarize several of PClean's built-in distributions here, before showing how to compose them into short PClean programs:

\begin{itemize}

\item \texttt{string\_prior(min, max)} encodes a prior over strings between $\textit{min}$ and $\textit{max}$ characters long. The length is uniformly distributed within that range, and characters follow a Markov model based on relative character bigram frequencies in English. 

\item \texttt{typos(str)} is a distribution over strings centered at $\texttt{str}$. The generative process it represents is to sample a number of typos from a negative binomial distribution whose number-of-trials parameter depends on the length of $\texttt{str}$. That many typos (random insertions, deletions, substitutions, or transpositions) are then performed. The likelihood is computed approximately using dynamic programming.

\item \texttt{maybe\_swap(x, ys, p)} returns a true value $\texttt{x}$ with probability $1-\texttt{p}$, but chooses a replacement uniformly from $\texttt{ys}$ otherwise.

\item \texttt{transformed\_normal(mean, std, bijection)} samples a real number from a Gaussian distribution with the given mean and standard deviation, but then applies a transformation (the bijection). We use this distribution to model unit errors.
\end{itemize}

\subsection{Discussion of expressiveness of PClean}

PClean imposes restrictions relative to universal PPLs, which helped us to develop an inference algorithm that, for many PClean programs, produces results quickly and scales to large datasets.
In this section, we discuss these restrictions and their implications for cleaning dirty data using PClean. 

{\bf Our non-parametric prior vs. explicit user-specified priors over number of objects and link structure.} A primary difference between general-purpose open-universe languages, like BLOG, and PClean's modeling language is that PClean does not give the user control over the prior distribution over the number of objects of each class, or which objects of particular classes are related to one another.\footnote{However, note that BLOG also has limitations when it comes to expressing priors over link structure. It allows users to specify predicates that the targets of a reference slot must satisfy, and the choice is then assumed to be uniform among all objects satisfying the predicate. Thus, BLOG cannot express that certain objects are more ``popular'' targets of reference slots than others\textemdash an assumption that is built in to PClean's Pitman-Yor-based model. We also note that by introducing additional classes, PClean can represent more interesting priors over link structure. For example, suppose $A.Y$ and $B.Y$ are two reference slots to objects from $C$, and we wish each reference slot to be filled using different distributions over the objects in $C$. We can create dummy classes for each reference slot, $AC$ and $BC$, each with a single reference slot ($AC.Y$ and $BC.Y$) to the target class $C$. We then have the reference slots $A.Y$ and $B.Y$ target $AC$ and $BC$ respectively, instead of directly targeting $C$. This implements a hierarchical Pitman-Yor process; by analogy with the HDP-LDA topic model, objects of $A$ and $B$ play the role of words from two different documents, and objects of class $C$ are the topics.} Instead, it imposes a domain-general non-parametric prior. This limitation might be mitigated by (1) the use of strength and discount hyperparameters of the Chinese Restaurant Process to control the prior expected size of each class (for a particular amount of data), and (2) the fact that in many data-cleaning applications, accurate prior knowledge about the number of objects may not be unavailable, or else is not a deciding factor in making cleaning judgments. 

Of course, there are exceptions. As an interesting example, consider the Hospital dataset: if we knew the population of each city, we may have been able to specify accurate priors over the number of distinct hospitals in each city, allowing us to resolve co-reference questions differently in small cities (where it is more likely that two hospitals reported with similar names are in fact the same hospital) and large cities (where it may be more plausible that two hospitals exist with very similar names). However, this factor is likely to be decisive only in high-uncertainty regimes (where the data entries themselves do not help much to resolve the co-reference question), and it is unclear whether a data-cleaning system should trust such high-uncertainty answers (vs. reporting `I don't know'\textemdash see Appendix~\ref{sec:uncertainty}). If the use case is such that it \textit{is} desirable to represent such priors, similar logic might be encoded in PClean by creating two different classes for hospitals in large and small cities, and allowing their strength and discount parameters to vary independently.

\textbf{On schemas with cyclic vs. acyclic class dependency graphs.} PClean requires that the schema of the latent database have an acyclic class dependency graph: there cannot be a chain of reference slots $K$ such that $T(C.K) = C$. Although, generally speaking, many relational modeling and inference tasks may be well-served by cyclic class dependencies, we found during literature review that none of the benchmark data-cleaning problems in \cite{Abedjan2016,Dallachiesat2013,Rekatsinas2017,Heidari2019,Hu2012,Mahdavi2019} were naturally modeled using cyclic class dependencies. In addition, \cite{Pasula2003,Milch2006}, who use BLOG for deduplication, do not use its support for reference cycles. There are, of course, some tasks for which cyclic references may be a natural fit, e.g. denoising genealogical data, where we may want to model that people have parents, who are other people, with many attributes inherited from one's ancestors. One could still model such datasets using coarser PClean models, e.g., by clustering people into families without modeling parent/child relationships explicitly. More generally, when we wish to model objects of the same class $C$ (e.g. \textit{Person}) as related via some chain of reference slots, we can often instead introduce an additional class $C'$ (e.g. \textit{Family}), and model any related objects of class $C$ as referring to a shared object of class $C'$.

\section{Additional Inference Details}
\label{sec:inference-details}

\subsection{Object-wise rejuvenation moves}
In sequential Monte Carlo, rejuvenation moves are transition kernels that preserve the current target distribution $\pi_i$, similar to the kernels used in Markov chain Monte Carlo algorithms. But we do not run them until convergence, instead using them to ``rejuvenate'' past decisions within SMC, in light of new data.

Any valid MCMC kernel for our model is also a valid rejuvenation kernel, and in particular, Gibbs kernels\textemdash which update a single variable in the latent state according to its full conditional distribution, keeping the rest of the state fixed\textemdash are a natural choice. However, variables in a model are often correlated, and it can be difficult to escape local modes by updating them one at a time. PClean uses \textit{object-wise blocked rejuvenation} to address this challenge. Object-wise rejuvenation moves update all attributes and reference slots of a single object $r$ in the latent database instance $\mathbf{R}$. In doing so, these moves may also lead to the ``garbage collection'' of objects that are no longer connected to the observed dataset, or to the insertion of new objects as targets of $r$'s reference slots. 

Let $r \in \mathbf{R}$ be any object in a relational database instance $\mathbf{R}$. Then we define $\mathbf{R}^{-r}$, $\mathbf{D}^{-r}$, $\Delta_r^{\mathbf{R}}$, $\mathbf{R}^r$, and $\mathbf{D}^r$ as follows:

\begin{itemize}
\item $\mathbf{R}^{-r}$ is the partial instance obtained by erasing from $\mathbf{R}$: (1) all attribute values and reference slot assignments for the object $r$; (2) all attribute values of objects $r'$ that depend on $r$; and (3) any objects $r'$ only accessible from $C_{obs}$ via slot chains that pass through $r$;

\item $\mathbf{D}^{-r}$ is the partial dataset obtained from $\mathbf{D}$ by erasing any attribute values whose distributions depend on values no longer specified within $\mathbf{R}^{-r}$;

\item $\Delta_r^{\mathbf{R}}$ is the partial instance specifying: (1) all attribute values and reference slot assignments for the object $r$; and (2) all objects $r'$ not in $\mathbf{R}^{-r}$ (accessible from $C_{obs}$ only via slot chains that pass through $r$), along with their attributes and reference slots;

\item $\mathbf{R}^r$ is the partial instance assigning values to all object attributes that depend on $r$'s attributes or reference slots as parents; and

\item $\mathbf{D}^r$ is the partial dataset assigning any attributes of observation objects that depend on on $r$'s attributes or reference slots as parents.
\end{itemize}

The model density then factorizes as:
$$p(\mathbf{R}, \mathbf{D}) = p(\mathbf{R}^{-r}, \mathbf{D}^{-r})p(\Delta_r^{\mathbf{R}} \mid \mathbf{R}^{-r})p(\mathbf{R}^r, \mathbf{D}^r \mid \Delta_r, \mathbf{R}^{-r}, \mathbf{D}^{-r}),$$

A \textit{blocked Gibbs sweep} loops through each object $r \in \mathbf{R}$ and updates it: $$\Delta_r^\mathbf{R} \sim p(\Delta_r^\mathbf{R} \mid \mathbf{R}^{-r}, \mathbf{D}, \mathbf{R}^r).$$
Because resimulating $\Delta_r^\mathbf{R}$ may delete objects from classes that are reachable from $r$ via reference slots, we perform this sweep in reverse topological order, starting with the objects that have no reference slots, and working our way up to the observation objects. If computing the blocked Gibbs distribution is intractable, then we can further divide $\Delta_r^\mathbf{R}$ according to user-specified subproblem decompositions for $\mathbf{Class}(r)$, as discussed in Section 3.3 of the paper. As the user subproblems get smaller in size, the algorithm approaches ordinary one-variable-at-a-time Gibbs sampling; thus, choosing subproblems is a simple way that users can trade off between runtime and accuracy, based both on the needs of their application and the specific properties of their models or datasets. 

Our rejuvenation kernels are compiled using PClean's proposal compiler, and as such, also benefit from (1) efficient enumeration strategies that take advantage of conditional independence in the variables being updated, and (2) user-specified `preferred values' inference hints (see Section 3.3). The paper's Algorithm 1 can be adapted for rejuvenation by adding observed variables to the Bayesian network for each attribute value specified in $\mathbf{R}^r$ (that is, each attribute value that, given the current link structure, depends on a latent variable being updated). Some of the variables within $\Delta_r^\mathbf{R}$ may be constrained by the observed dataset $\mathbf{D}$; this will depend on the patterns of missingness in the observations that, under the current link structure, are connected in some way to the object being updated. PClean recognizes when these patterns of missingness change (due to link structure changing), and compiles new proposals as necessary.

\subsection{Continuous variables and parameters}

PClean allows users to include continuous variables in their models, either as parameters or attributes in class declarations. To handle these, we augment the inference algorithm in three additional ways:

\begin{enumerate}
	\item \textbf{Gibbs rejuvenation for parameter values.} Continuous parameters $\theta$ are updated during SMC via separate Gibbs rejuvenation moves. PClean recognizes certain conjugate relationships between parameter hyperpriors and the attribute statements that use the parameters (e.g., Normal/Normal, Beta/Bernoulli, and Dirichlet/Categorical), and automatically exploits these for efficient and rejuvenation moves informed by all the relevant data. The inference engine tracks the relevant sufficient statistics as inference progresses, so these updates need not perform costly counts or summations.

	\item \textbf{Mixing with the prior for proposals of continuous attributes.} Continuous attributes are handled as though they are discrete variables with `preferred values' set to $\emptyset$. The effect of this is that the locally optimal proposal for discrete variables is first derived without regard for the latent continuous attributes being proposed as part of the same subproblem (meaning that any likelihoods that depend on latent continuous attributes are not included during enumeration); then, once discrete values have been sampled, continuous values are sampled from their prior CPDs given any of their parent values (which may have been more intelligently proposed).

	\item \textbf{Particle Gibbs object-wise rejuvenation.} Because the proposals generated by technique (2) for continuous variables may be poor, Metropolis-Hastings may often reject. To improve chances of acceptance, users can enable Particle Gibbs rejuvenation, which, in order to propose an update $\Delta_r^\mathbf{R}$ to an object $r$ of class $C$, runs \textit{conditional SMC} on the sequence of user-defined subproblems within class $C$. Using Particle Gibbs, PClean can compensate for poorer proposals by sampling many weighted particles for each subproblem, which are combined into a joint proposal for the object. Note that without continuous variables, Metropolis-Hastings is generally preferred.
\end{enumerate}

\subsection{Optimality conditions for proposal compiler}

The proposal compiler produces smart proposals by efficiently enumerating discrete variables (exploiting conditional independence) and computing only those likelihood terms that are necessary for a particular SMC or MCMC update. When all latent variables within a subproblem have finite discrete domains, and no variables have preferred values hints specified, the proposals PClean produces are locally optimal SMC proposals, as defined in~\cite{naesseth2019elements}, or, for MCMC, exact blocked Gibbs rejuvenation kernels. However, introducing preferred-values hints that do not completely cover the posterior mass, or using continuous attributes within the subproblem, will lead to suboptimal (but faster-to-compute) proposals.

\subsection{Observation hashing}

Preferred values hints can help to limit the number of possibilities enumeration must consider for attribute values, but reference slots can also pose a problem: as the sequential Monte Carlo algorithm progresses, the latent database fills up with objects that could serve as possible targets, and considering each of them can be expensive.

In many models, however, the value of a reference slot is highly constrained by observations in $\mathbf{D}$. Consider an object $r$ of class $C$ with reference slot $Y$, and let $\mathcal{W} = \{W \mid T(C_{obs}.W) = C\}$ be the set of slot chains connecting observation objects to objects of class $C$. Given a query map $\mathbf{Q}$, we can check if there exist any observed attributes $x \in \mathcal{A}(\mathbf{D})$ that $\mathbf{Q}$ maps to a slot chain beginning $W.Y$. For each $W \in \mathcal{W}$, let $\mathcal{K}_{W, C.Y} = \{(U, x) \mid x \in \mathcal{A}(\mathbf{D}) \wedge \mathbf{Q}(x) = W.Y.U \}$. Then the only objects of the target class $T(C.Y)$ that $r.Y$ can possibly point to are $$\bigcap_{W \in \mathcal{W}} \bigcap_{\{i \textrm{ s.t. } r^{obs}_i.W = r\}} \bigcap_{(U, x) \in \mathcal{K}_W} \{r' \in \mathbf{R}_{T(C.Y)} \mid d_i.x = r'.U\}.$$
PClean can maintain, for each class, an index that maps values $v_x$ to sets of objects $r'$ such that $r'.U = v_x$. PClean also maintains back-pointers from objects $r$ to the observation objects that reference them, and stores with each object $r$ the observed attribute values $d_i.x$ that constrain it. This allows PClean to compute the set of legal target objects for a given reference slot in $O(|\mathcal{W}|)$ time, which is constant in the number of latent objects for many models. (Indexing does require memory. Users can optionally control which $U$ values are indexed on by including $\textbf{index on } U$ statements within class declarations.) Of course, in some models and datasets, the size of the computed set of possible target objects may still be large, necessitating enumeration. But in common cases where the vast majority of possible targets have zero likelihood, this indexing plays a key role in helping PClean to scale to large datasets.



\end{document}